\documentclass{article}

\usepackage{preprint}

\usepackage{rotating}
\usepackage{floatrow}
\usepackage{multirow}
\usepackage{graphicx}
\usepackage{soul}
\usepackage{comment}
\usepackage{multicol}
\usepackage{comment}
\usepackage{amsmath}
\usepackage{cite}
\usepackage{booktabs, makecell, tabularx}
\usepackage{amsfonts}
\usepackage[dvipsnames,table,xcdraw]{xcolor}
\definecolor{gold}{rgb}{0.85, 0.65, 0.13}
\usepackage{algorithm}
\usepackage[noend]{algpseudocode}

\algnewcommand\algorithmicforeach{\textbf{for each}}
\algdef{S}[FOR]{ForEach}[1]{\algorithmicforeach\ #1\ \algorithmicdo}
\algnewcommand{\algorithmicand}{\textbf{ and }}
\algnewcommand{\algorithmicor}{\textbf{ or }}
\algnewcommand{\OR}{\algorithmicor}
\algnewcommand{\AND}{\algorithmicand}
\algnewcommand{\LineComment}[1]{\State \(\triangleright\) {\footnotesize #1}}
\newcommand{\pluseq}{\mathrel{+}=}
\algdef{SE}[VARIABLES]{Variables}{EndVariables}
   {\algorithmicvariables}
   {\algorithmicend\ \algorithmicvariables}
\algnewcommand{\algorithmicvariables}{\textbf{global variables}}

\usepackage[caption=false]{subfig}
\usepackage{pgfplots}

\pgfplotsset{compat=1.12}

\newcommand{\ie}{{\it i.e.}}

\title{Continuous Ant-Based Neural Topology Search}

\author{
  AbdElRahman A. ElSaid\\
  \texttt{aelsaid@mail.rit.edu} \\
  \And
  Joshua Karns\\
  \texttt{josh@mail.rit.edu} \\
  \And
  Zimeng Lyu\\
  \texttt{zimenglyu@mail.rit.edu} \\
  
  \par
  \And
  Alexander G. Ororbia \\
  \texttt{ago@cs.rit.edu} \\
  \\
  \And
  Travis J. Desell\\
  \texttt{tjdvse@rit.edu} \\
  
  \\
    Golisano College of Computing and Information Sciences\\
    Rochester Institute of Technology\\
    Rochester, NY 14623\\
}

\begin{document}
\maketitle

\begin{abstract}
This work introduces a novel, nature-inspired neural architecture search (NAS) algorithm based on ant colony optimization, Continuous Ant-based Neural Topology Search (CANTS), which utilizes synthetic ants that move over a continuous search space based on the density and distribution of pheromones, is strongly inspired by how ants move in the real world. The paths taken by the ant agents through the search space are utilized to construct artificial neural networks (ANNs). This continuous search space allows CANTS to automate the design of ANNs of any size, removing a key limitation inherent to many current NAS algorithms that must operate within structures with a size predetermined by the user. CANTS employs a distributed asynchronous strategy which allows it to scale to large-scale high performance computing resources, works with a variety of recurrent memory cell structures, and makes use of a communal weight sharing strategy to reduce training time. The proposed procedure is evaluated on three real-world, time series prediction problems in the field of power systems and compared to two state-of-the-art algorithms. Results show that CANTS is able to provide improved or competitive results on all of these problems, while also being easier to use, requiring half the number of user-specified hyper-parameters.
\end{abstract}

\section{Introduction}
\label{sec:introduction}

Manually optimizing artificial neural network (ANN) structures has been an obstacle to the advancement of machine learning as it is significantly time-consuming and requires a considerable level of domain expertise~\cite{zoph2016neural}. The structure of an ANN is typically chosen based on its reputation based on results of existent literature or based on knowledge shared across the machine learning community, however changing even a few problem-specific meta-parameters can lead to poor generalization upon committing to a specific topology~\cite{erkaymaz2014impact,Barna1990}.  To address these challenges, a number of neural architecture search (NAS)~\cite{elsken2018neural,liu2018darts,pham2018efficient,xie2018snas,luo2018neural,zoph2016neural} and neuroevolution (NE)~\cite{stanley2019designing,darwish2020survey} algorithms have been developed to automate the process of ANN design. More recently, nature-inspired neural architecture search (NI-NAS) algorithms have shown increasing promise, including the Artificial Bee Colony (ABC) optimization procedure~\cite{horng2017fine}, the Bat algorithm~\cite{yang2010new}, the Firefly algorithm~\cite{yang2010nature}, and the Cuckoo Search algorithm \cite{leke2017deep}.

Among the more recently successful applied NI-NAS strategies are those based on ant colony optimization (ACO)~\cite{dorigo1996ant}, which have proven to be particularly powerful when automating the  design of recurrent neural networks (RNNs). Originally, ACO for NAS was limited to small structures based on Jordan and Elman RNNs~\cite{desell2015evolving} or was used as a process for reducing the number of network inputs~\cite{mavrovouniotis2013evolving}. Later work proposed generalizations of ACO for optimizing the synaptic connections of RNN memory cell structures~\cite{elsaid2018optimizing} and even entire RNN architectures in an algorithmic framework called Ant-based Neural Topology Search (ANTS)~\cite{elsaid2020ant}. In the ANTS process, ants traverse a single massively-connected ``superstructure'',  which contains all of the possible ways that the nodes of an RNN may connect with each other, both in terms of structure (\ie, all possible feed forward connections), and in time (\ie, all possible recurrent synapses that span many different time delays), searching for optimal RNN sub-networks. This approach shares similarity to NAS methods in ANN cell and architecture design~\cite{liu2018darts,pham2018efficient,xie2018snas,cai2018proxylessnas,
guo2020single,bender2018understanding,dong2019one,zhao2020few}, which operate within a limited search space, generating cells or architectures with a pre-determined number of nodes and edges~\cite{elsken2018neural}, 

Most NE methods, instead of operating within fixed bounds, are constructive (e.g. NEAT~\cite{stanley2002evolving} and EXAMM~\cite{ororbia2019examm}); they start with a minimum configuration for the ANN and then add or remove elements over several iterations of an evolutionary process. More advanced strategies involve generative encoding, such as HyperNEAT~\cite{stanley2009hypercube}, where a generative network is evolved, which can then be used to create architectures and assign values to their synaptic weights. Nonetheless, these approaches still require manually specifying or constraining the size or scale of the generated architecture in terms of the number of layers and nodes. 

Constructive NAS methods often suffer from getting stuck in (early) local minima or take considerable computation time to evolve structures that are sufficiently large in order to effectively address the task at hand, especially for large-scale deep learning problems. Alternately, having to pre-specify bounds for the space of possible NAS-selected architectures can lead to poorly performing or suboptimal networks if the bounds are incorrect, requiring many runs of varying bound values. In order to address these challenges, this work introduces the novel ACO-inspired algorithm, \emph{Continuous Ant-based Neural Topology Search (CANTS)}, which utilizes a continuous search domain that flexibly allows for the design of ANNs of any size. Synthetic continuous ant (\emph{cant}) agents move through this search space based on the density and distribution of pheromone signals, which emulates how ants swarm in the real world, and the paths resulting from their exploration are used to construct RNN architectures. CANTS is a distributed, asynchronous algorithm, which facilitates scalable usage of high performance computing (HPC) resources, and also utilizes communal intelligence to reduce the amount of training required for candidate evolved networks. The procedure further allows for the selection of recurrent nodes from a suite of simple neurons and complex memory cells used in modern RNNs: $\Delta$-RNN units~\cite{ororbia2017diff}, gated recurrent units (GRUs)~\cite{chung2014empirical}, long short-term memory cells (LSTMs)~\cite{hochreiter1997long}, minimal gated units (MGUs)~\cite{zhou2016minimal}, and update-gate RNN cells (UGRNNs)~\cite{collins2016capacity}. 

In this work, CANTS is compared to state-of-the-art benchmark algorithms used in designing RNNs for time series data prediction: ANTS~\cite{elsaid2020ant} and EXAMM~\cite{ororbia2019examm}. In addition to eliminating the requirement for pre-specified architecture bounds, CANTS is shown to yield results that improve upon or are competitive to ANTS and EXAMM while reducing the number of user specified hyperparameters from $16$ in EXAMM and $16$ in ANTS down to $8$ in CANTS. CANTS also provides an advancement to the field of ant colony optimization, as it is the first algorithm capable of optimizing complex graph structures without requiring a predefined superstructure to operate within. While ACO has been applied to continuous domain problems before~\cite{socha2008ant,kuhn2002ant,xiao2011hybrid,gupta2014transistor,bilchev1995ant}, to the authors' knowledge, our algorithm is the first to utilize simulate the movements of ants through a continuous space to design unbounded graph structures.

\section{Methodology}
\label{sec:method}

\begin{algorithm}
	\footnotesize{}
    \caption{Continuous Ant-guided Neural Topology Search Algorithm}\label{alg:cants_pseudo}
    \begin{algorithmic}
		\Procedure{$Work Generator$}{}
			\LineComment{Construct search space with inputs at y=0 and output at y=1}
			\LineComment{Recurrent time steps is the spaces's z axis}
			\State $search\_space = \textbf{new}~SearchSpace$
			\For {$i \gets 1 \dots max\_iteration$}
				\State $nn_{new} \gets AntsSwarm( )$
				\State $send\_to\_worker (nn_{new}, worker.id)$
				\State $nn_{new}, fit \gets receive\_fit\_from\_worker( )$
				\If { $nn\_{fitness}< worst\_population\_member$ }
					\State $population.pop ( worst\_population\_member)$
					\State $population.add ( nn_{new} )$
					\State $RewardPoints ( nn_{new} )$
				\EndIf			
			\EndFor
		\EndProcedure
		
		\Procedure{$Worker$}{}
			\State $receive\_from\_master (nn)$		 
			\State $fitness \gets train\_test\_nn ( nn )$
			\State $send\_fitness\_to\_master ( nn, fitness )$
		\EndProcedure

		\Procedure {$AntsSwarm$} {}
			\LineComment {Ants choose input in discrete fashion}
			\For {$ant \gets 1 \dots no\_ants$}
				\State $CreatePath( ant )$
			\EndFor
			\LineComment { Use DBscan to cluster ants paths points }
			\State { $segments \gets DBscanPaths(ants)$ }
			\LineComment {Create RNN from segments}
			\State $rnn_{new} \gets CreateRNN(segments)$
			\Return $rnn_{new}$
		\EndProcedure

		\Procedure {$CreatePath$} {$ant$}
			\LineComment {Choose input in discrete fashion}
			\State $ChooseInput( ant )$
			\LineComment { Create a path starting from the input  }
			\While { $ant.current\_y<0.99$}
				\State $r \gets \textbf{uniform\_random}(0, pheromone\_sum - 1)$
				\State $ant.current\_level  \gets ant.climb$
				\If { $r > ant.exploration\_instinct $ or $ search\_space[ant.current\_level] $ is not $ Empty  $ } 
					\State $point \gets CreateNewPoint(ant.search\_radius)$
					\State $ant.path.insert( point )$
					\State $search\_space.insert( point )$
				\Else
					\State $point \gets FindCenterOfMass(ant.current\_position, ant.search\_radius)$
					\If { $point$ not in $search\_space[ant.level]$}
						\State $ant\_path.insert( point )$
					\EndIf
				\EndIf
			\EndWhile
			\LineComment {Choose Output in discrete fashion}
			\State $ChooseOutput(ant)$
		\EndProcedure

		        \algstore{myalg}
		    \end{algorithmic}
		\end{algorithm}

		\begin{algorithm}
		 		    \begin{algorithmic}
		 		        \algrestore{myalg}

		\Procedure {$ChooseInput$} {$ant$} 
		    \LineComment{Select input probabilistically according to pheromones}
			\State $pheromone\_sum \gets \textbf{sum}(pheromones.input)$
			\State $r \gets \textbf{uniform\_random}(0, pheromone\_sum - 1)$
			\State $ant.input \gets 0$
			\While {$r > 0$}:
	            \If {$r < pheromones.input[ant.input]$}
	                \State $ant.input \gets 1$
	                \State \textbf{break}
	            \Else
	                \State $r \gets r - pheromones.input[ant.input]$
	                \State $ant.input \gets ant.input + 1$
	            \EndIf
			\EndWhile
		\EndProcedure

		\Procedure {$ChooseOutput$} {$ant$} 
		    \LineComment{Select input probabilistically according to pheromones}
			\State $pheromone\_sum \gets \textbf{sum}(pheromones.output)$
			\State $r \gets \textbf{uniform\_random}(0, pheromone\_sum - 1)$
			\State $ant.input \gets 0$
			\While {$r > 0$}:
	            \If {$r < pheromones.input[ant.output]$}
	                \State $ant.output \gets 1$
	                \State \textbf{break}
	            \Else
	                \State $r \gets r - pheromones.output[ant.output]$
	                \State $ant.output \gets ant.output + 1$
	            \EndIf
			\EndWhile
		\EndProcedure

		\Procedure {$DBscanPaths$} {$ants$} 
			\For {$ant \gets 1 \dots num\_ants$}
				\For {$point \gets 1 \dots ant\_path$}
					\State {$segments[ant].insert(PickPoint(point))$}
				\EndFor
			\EndFor
			\Return $segments$
		\EndProcedure
		
		\Procedure { $PickPoint$} {$point$}
			\State $ [node, points_cluster] \gets DBscane(point, search\_space[point.level])$
			\State $node.out\_edges\_weights.insert(AvrgWeights(points\_cluster))$
			\State $search\_space.insert(node)$
			\Return $node$
		\EndProcedure

		\Procedure {$RewardPoints$} {$rnn$} 
			\ForEach {$node \in rnn.nodes$} 
				\State {$search\_space[ node ].pheromone \pluseq constant $}
				\State {$search\_space[ node ].weight \gets average_weight(node.weight, search\_space[ node ].weight)$}
				\If { $search\_space[ node ].pheromone > PHEROMONE\_THRESHOLD$ }
					\State { $search\_space[ node ].pheromone = PHEROMONE\_THRESHOLD$ }
				\EndIf 
			\EndFor
		\EndProcedure
	
\end{algorithmic}
\end{algorithm}

The CANTS procedure (see high-level pseudo-code in Algorithm~\ref{alg:cants_pseudo}) employs an asynchronous, distributed ``work-stealing'' strategy to allow for scalable execution on HPC systems. The work generation process maintains a population of the best-found RNN architectures and repeatedly generates candidate RNNs whenever the worker processes request them. This strategy allows workers to complete the training of the generated RNNs at whatever speed they are capable of, yielding an algorithm that is naturally load-balanced.  Unlike synchronous parallel evolutionary strategies, CANTS scales up to any number of available processors, supporting population sizes that are independent of processor availability. When the resulting fitness of candidate RNNs are reported to the work generator process, \ie, mean squared error over validation data, if the candidate RNN is better than the worst RNN in the population, then the worst RNN is removed and the candidate is added. Note that the saved pheromone placement points for the candidate are incremented in the continuous search space.

\begin{figure}
    \centering
    \begin{tabular}{cc}
    \subfloat[Initially generated RNN\label{fig:frame1}]{
        \centering
        \includegraphics[width=.49\textwidth]{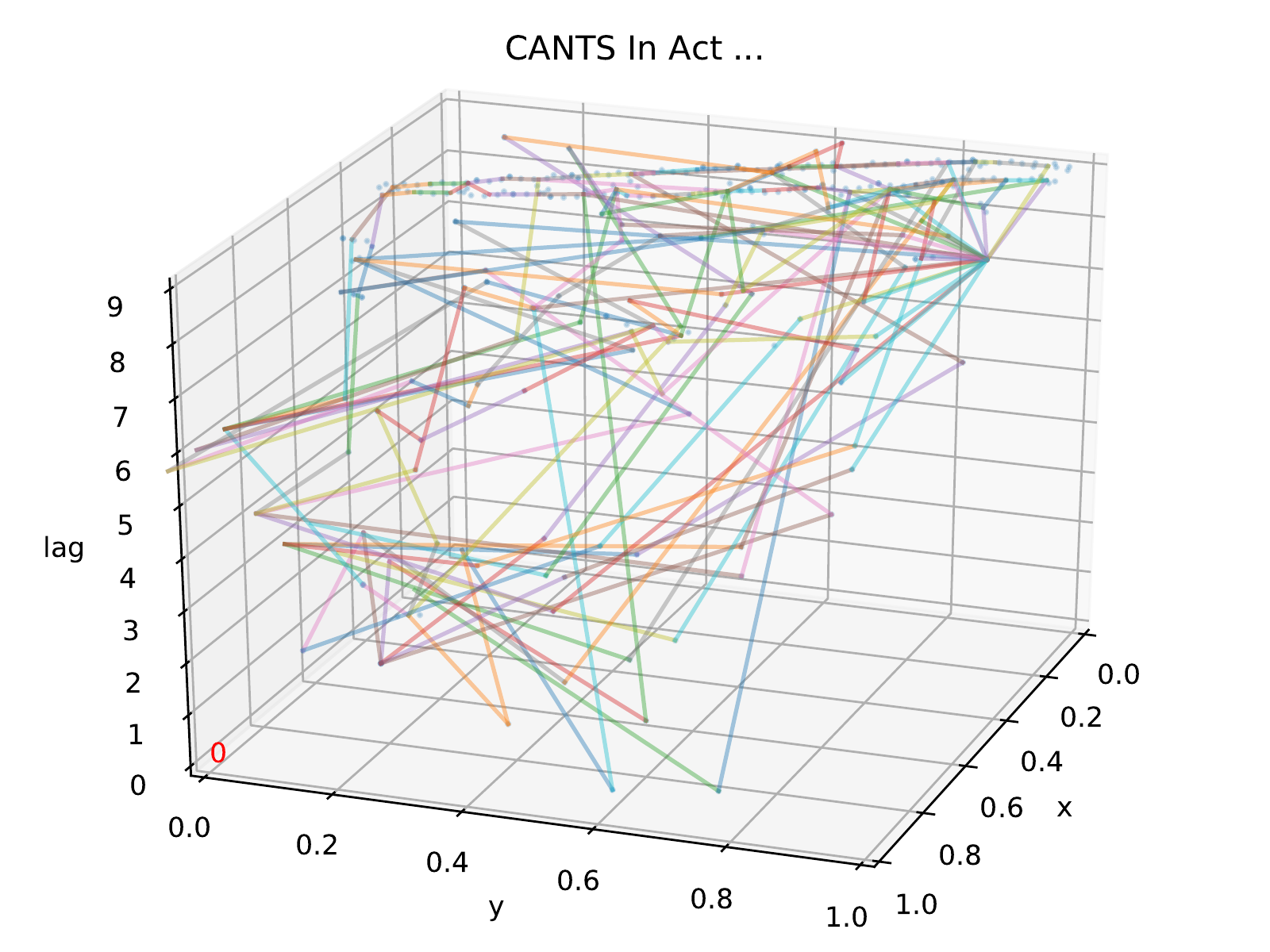}
    } &
    \subfloat[123rd generated RNN\label{fig:frame2}]{
        \centering
        \includegraphics[width=.49\textwidth]{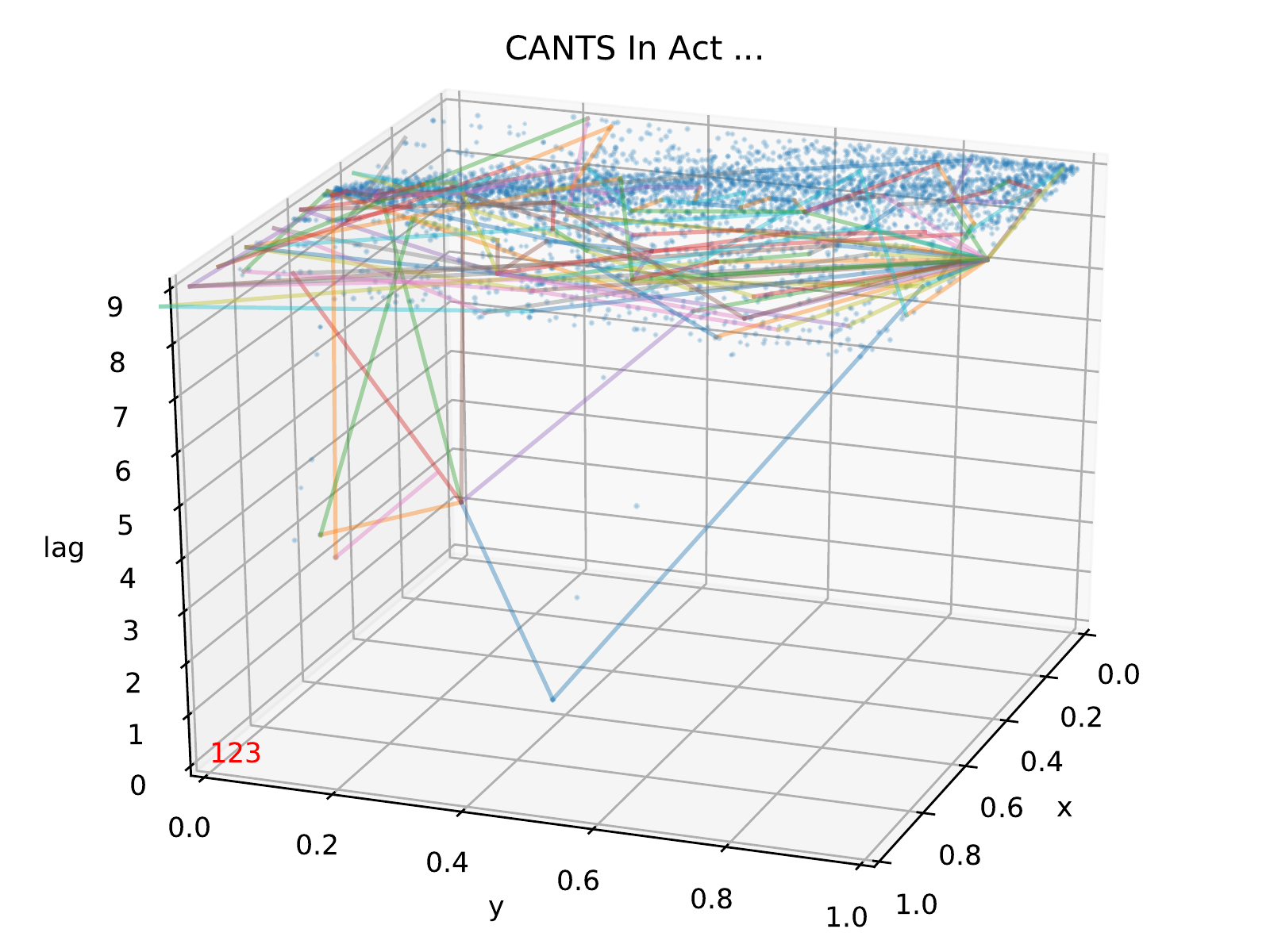}
    }
    \caption{A replay visualization of CANTS showing the pheromones and paths taken by ants through the complex, continuous search space.}
    \label{fig:cants_replay_visualization}
    \end{tabular}
\end{figure}

Candidate RNNs are synthesized using a search space that could be likened to a stack of continuous 2D planes, where each 2D plane or slice of this stack represents a particular time step $t$ (see Figure \ref{fig:cants_move_explore}).
The input nodes for each time step are uniformly distributed at the input edge of the search space. A synthetic continuous ant agent (or \emph{cant}) picks one of the discrete input node positions to start at and then moves through the continuous space based on the current density and distribution of other pheromone placements. Cants are allowed to move forward on the level they are on and can move up to any of the ones above it. However, they are restricted from moving down the stack -- this constraint is imposed because the movement of ants between the layers in our algorithm's search space represents the forward propagation across time-steps, hence it is only possible to propagate information from a previous time step ($t-k$) up to and including the current step $t$ but not the reverse, since this would imply carrying unknown, future signals backwards. While ants only move forward on a given plane, they are permitted to move backward on a slice if they had just moved to it form a lower level (since many RNNs have synapses that potentially skip neuronal layers). This enforced upward and (overall) forward movement ensures that cants continue to progress towards outputs and do not needlessly circle around in the search space. Figures~\ref{fig:cants_move} shows examples of how cants move from an input edge of the search space to its output edge, how cants explore new regions in the search space, how cants exploit previously searched areas via attraction to deposited pheromones, and how cant path through the space are translated into a final, candidate RNN.  The software developed that implements our CANTS procedure also provides a replay visualization tool so that traces from a run can be visualized to see how pheromones are deposited and how RNNs are generated, as shown in Figure~\ref{fig:cants_replay_visualization}.

\begin{figure}
    \centering 
    \begin{tabular}{cc}
        \subfloat[][]{\includegraphics[width=0.475\textwidth, height=0.15\textheight]{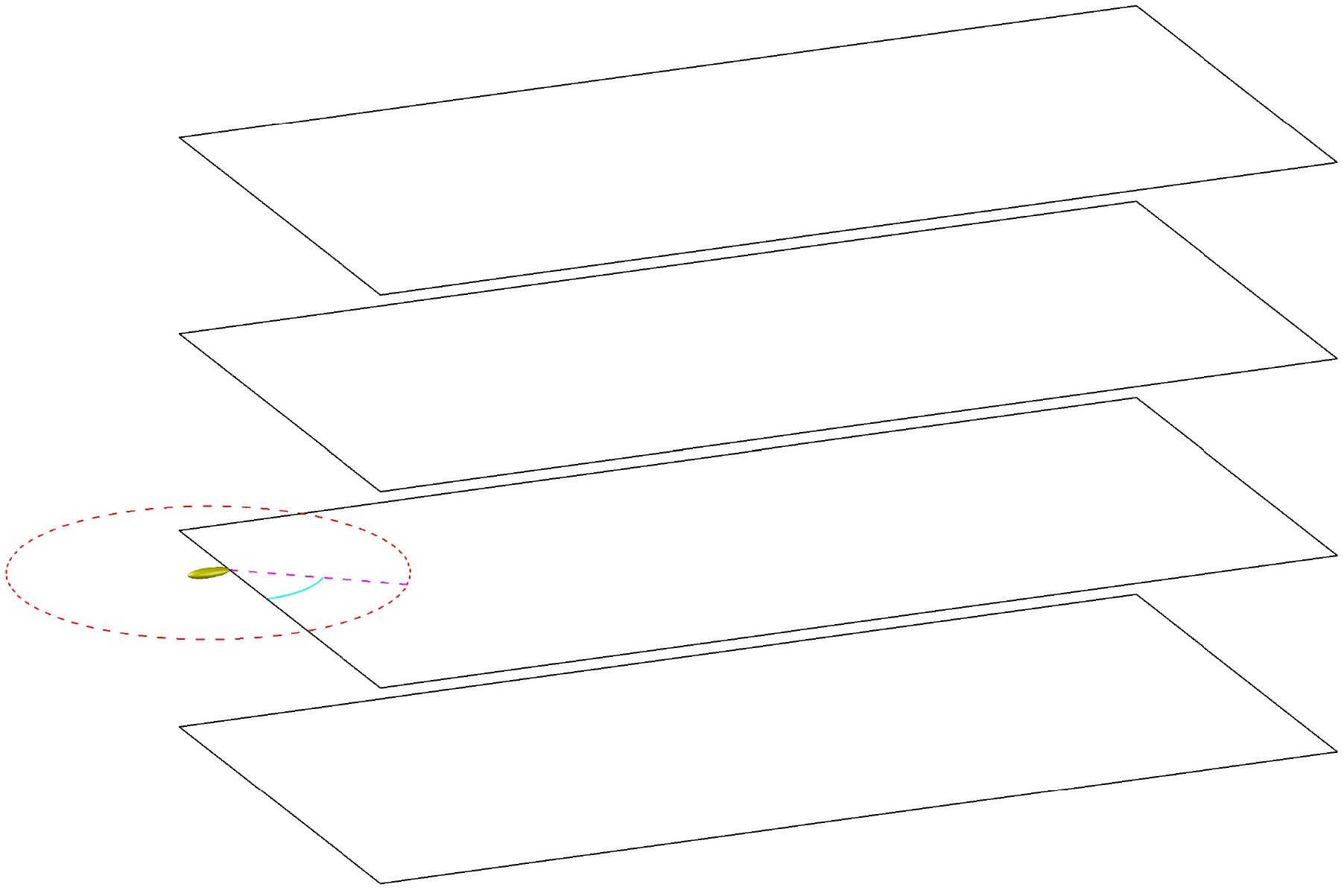} \label{fig:cants_move_explore}} &
        \subfloat[][]{\includegraphics[width=0.475\textwidth, height=0.15\textheight]{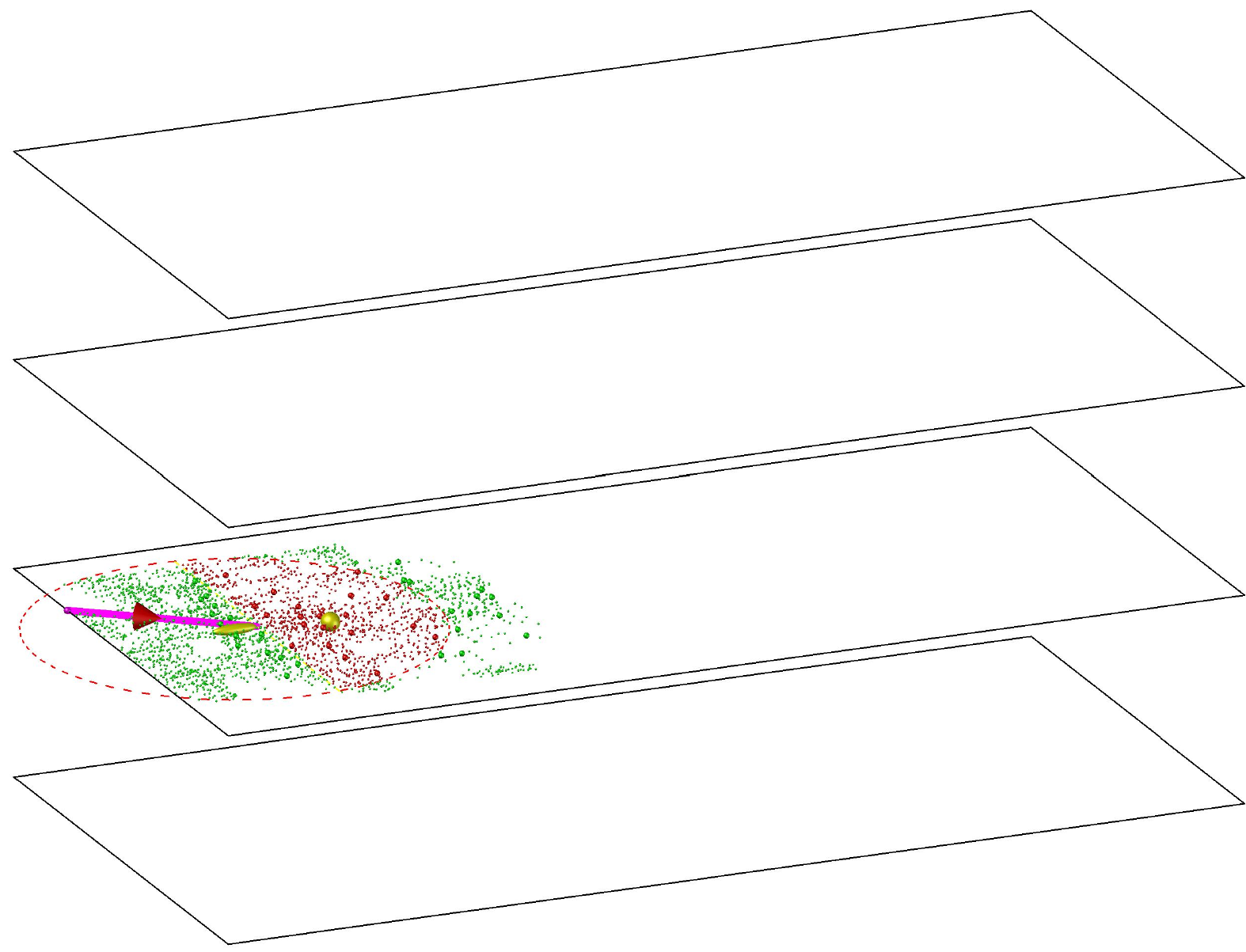} \label{fig:cants_move_exploit_1}} \\

        \subfloat[][]{\includegraphics[width=0.475\textwidth, height=0.15\textheight]{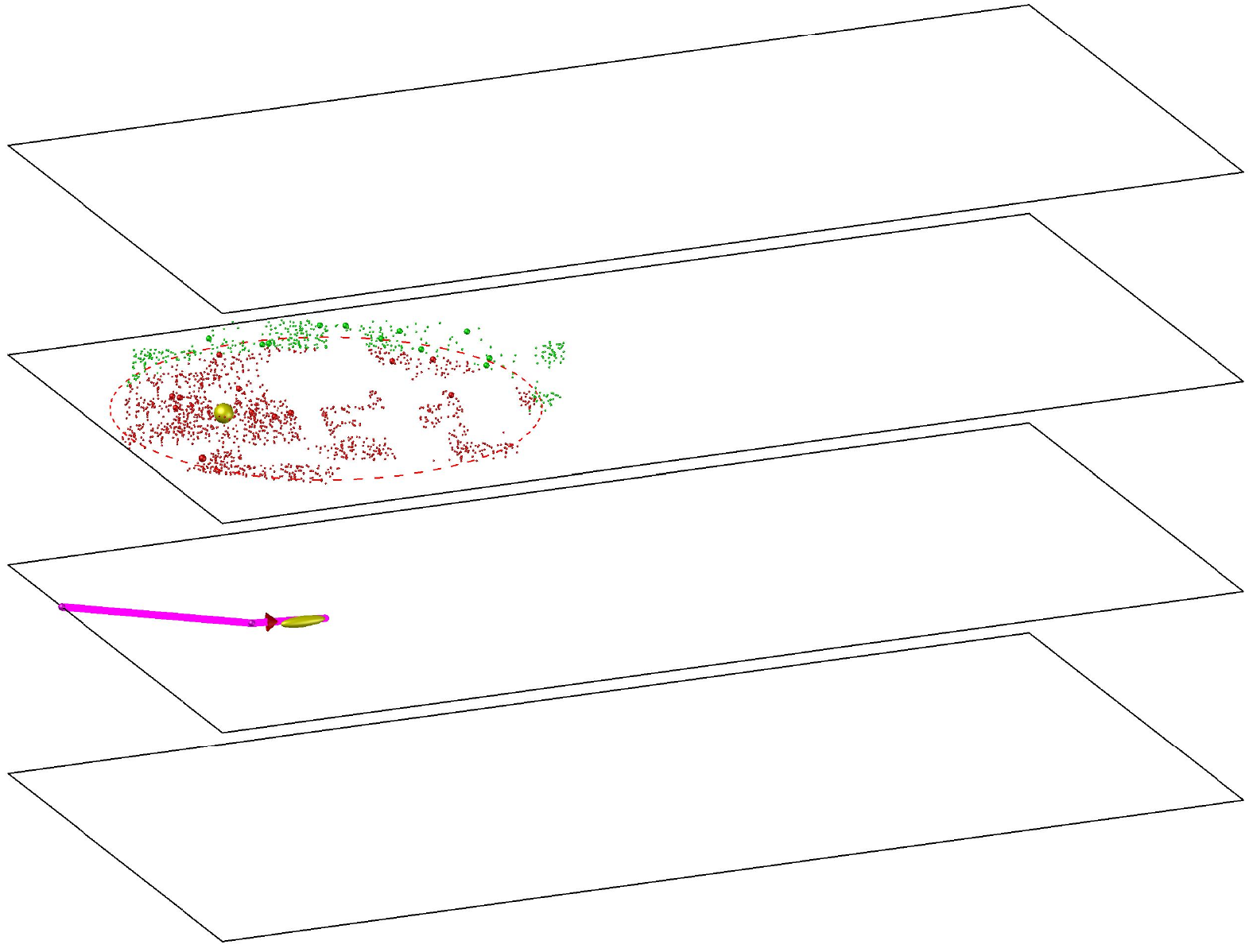} \label{fig:cants_move_exploit_2}} &
        \subfloat[][]{\includegraphics[width=0.475\textwidth, height=0.15\textheight]{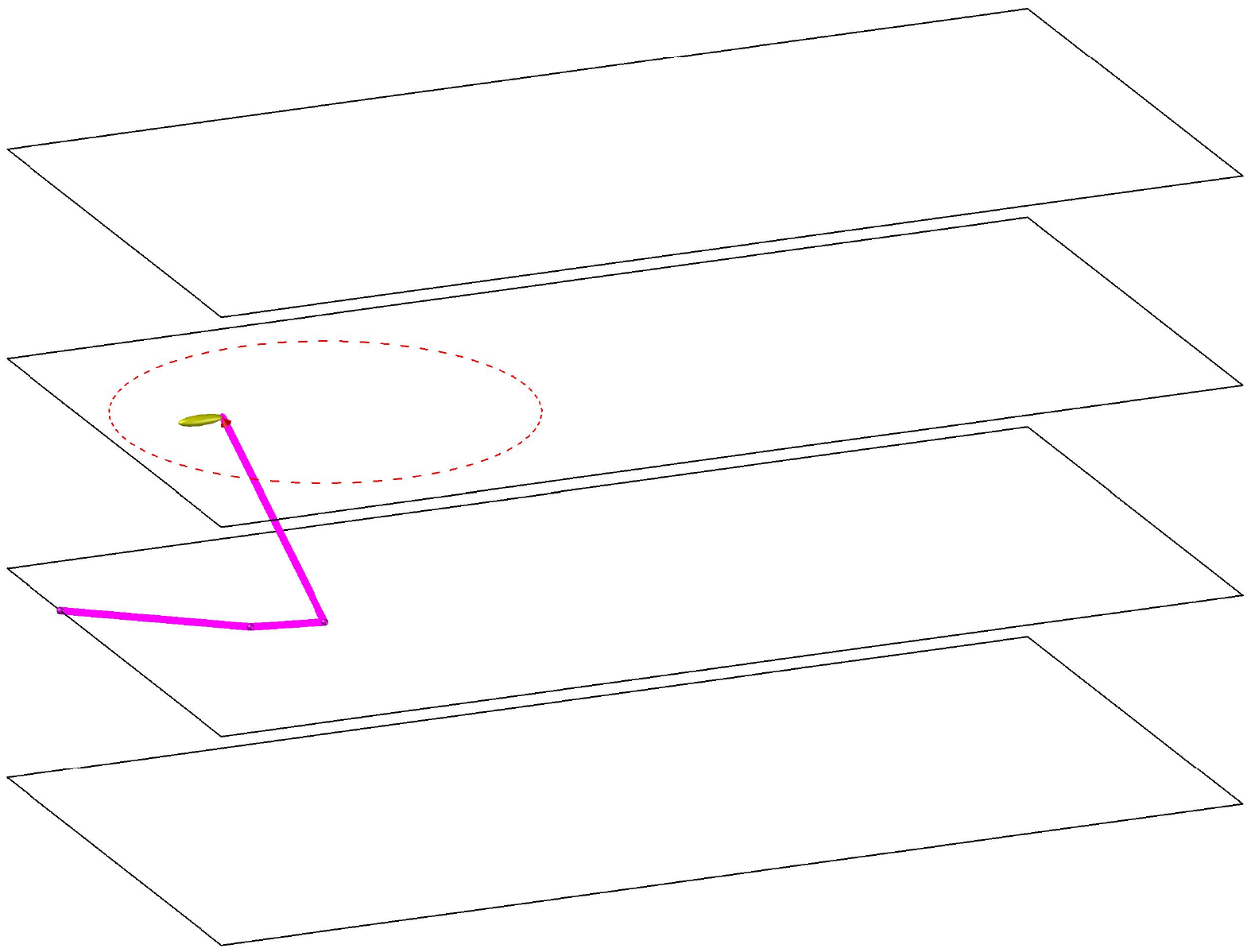} \label{fig:cants_move_05}} \\
        
        \subfloat[][]{\includegraphics[width=0.475\textwidth, height=0.15\textheight]{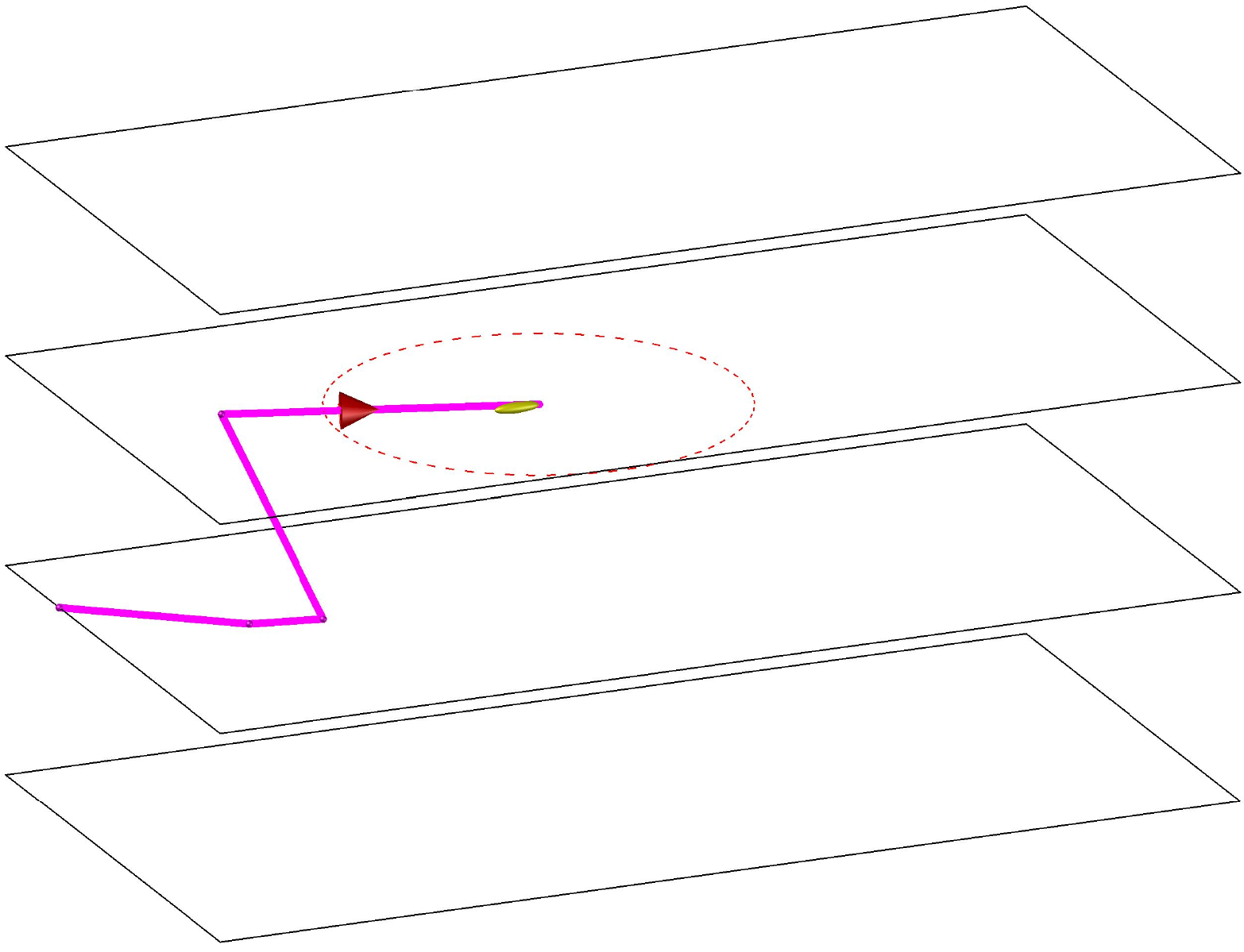} \label{fig:cants_move_06}} 
        &
        \subfloat[][]{\includegraphics[width=0.475\textwidth,height=0.15\textheight]{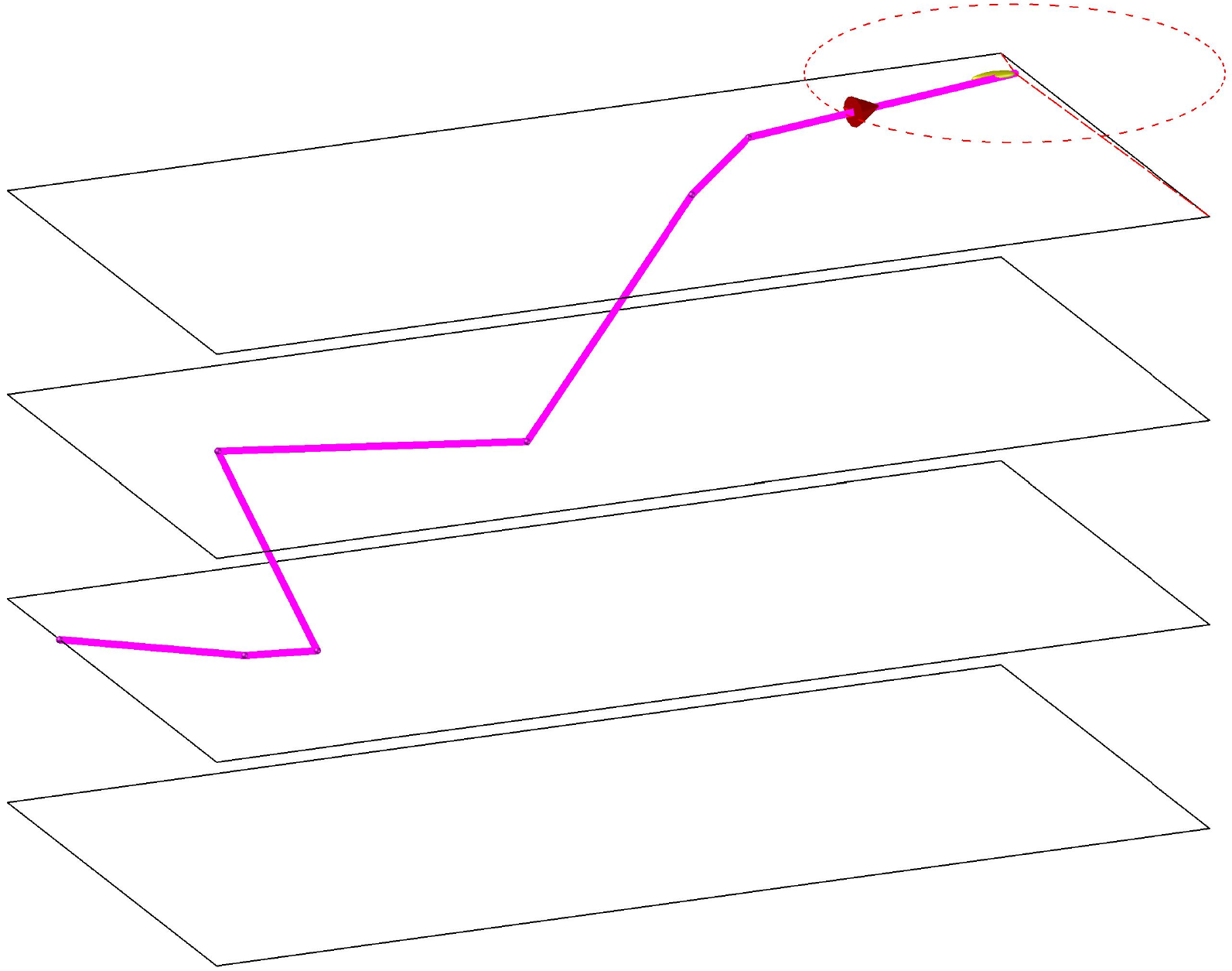} \label{fig:cants_move_pick_output}} 
        \\

    \end{tabular}
    \caption[]{ {\bf Cant path selection and network construction:} (a) After an cant picks a layer to start with and an input node, it decides if it will move to a new random point (exploration), or follow pheromone traces (exploitation). If the the former, the cant will randomly pick a forward {\color{cyan}{angle}} between $0^\circ$ and $180^\circ$ and move in that direction equal to its its {\color{red}sensing radius}. (b) When the cant wants to use pheromone traces to determine its new point, it will first sense the the pheromone traces within its sensing radius. The example cant did not change its layer, so the cant will only consider the pheromone traces in front of it and not move backwards. The ant will then calculate the center of mass of the pheromone traces within its sensing radius and then move to the center of their mass ({\color{gold}sphere}). (c) When the cant moves to a level above it and decides that it will use exploitation, it will consider the pheromone traces in its sensing range in all directions, which lie between the angles $0^\circ$ and $360^\circ$. This way, the cant can move backwards when jumping from a layer to another which makes a recurrent connection that goes back between hidden layers. (d) The cant moves upward to the higher level. (e) The cant will moves to a new point by exploration. (f) After a series of upward and forward moves by either exploration or exploitation, when the cant has output nodes within its sensing radius, it will stop the continuous search and select picking an output node based on their discrete pheromone values. If there is only one output node, then the cant will directly connect its last point to the output.
    \label{fig:cants_move_1}}
\end{figure}

\begin{figure}
    \ContinuedFloat 
    \centering 
    \begin{tabular}{cc}
        \subfloat[][\label{fig:ants_multi_path}]{\includegraphics[width=0.475\textwidth,height=0.25\textheight]{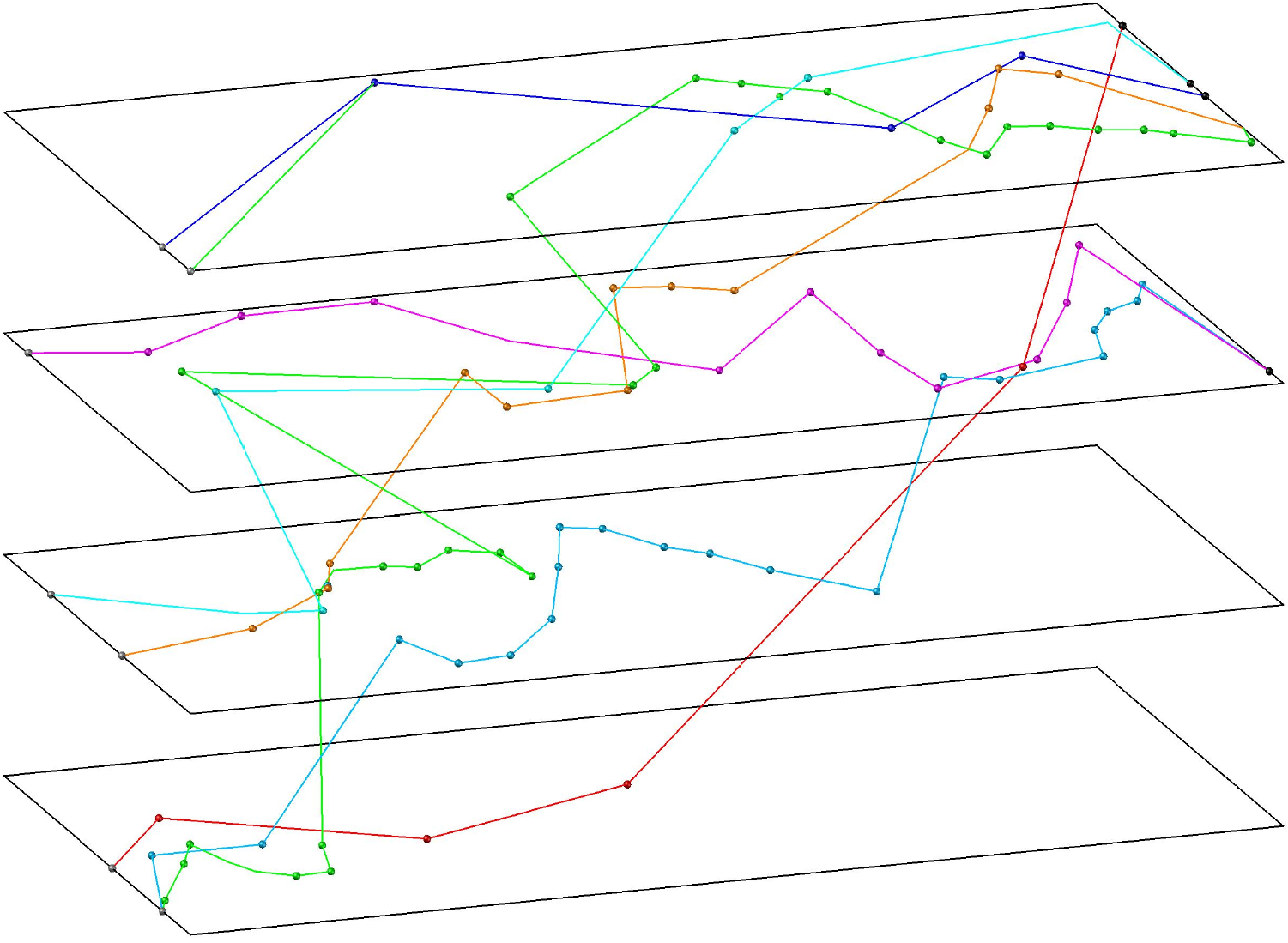} \label{fig:cants_move_10}} 
        &
        \subfloat[][\label{fig:ants_condensed_path}]{\includegraphics[width=0.45\textwidth, height=0.25\textheight]{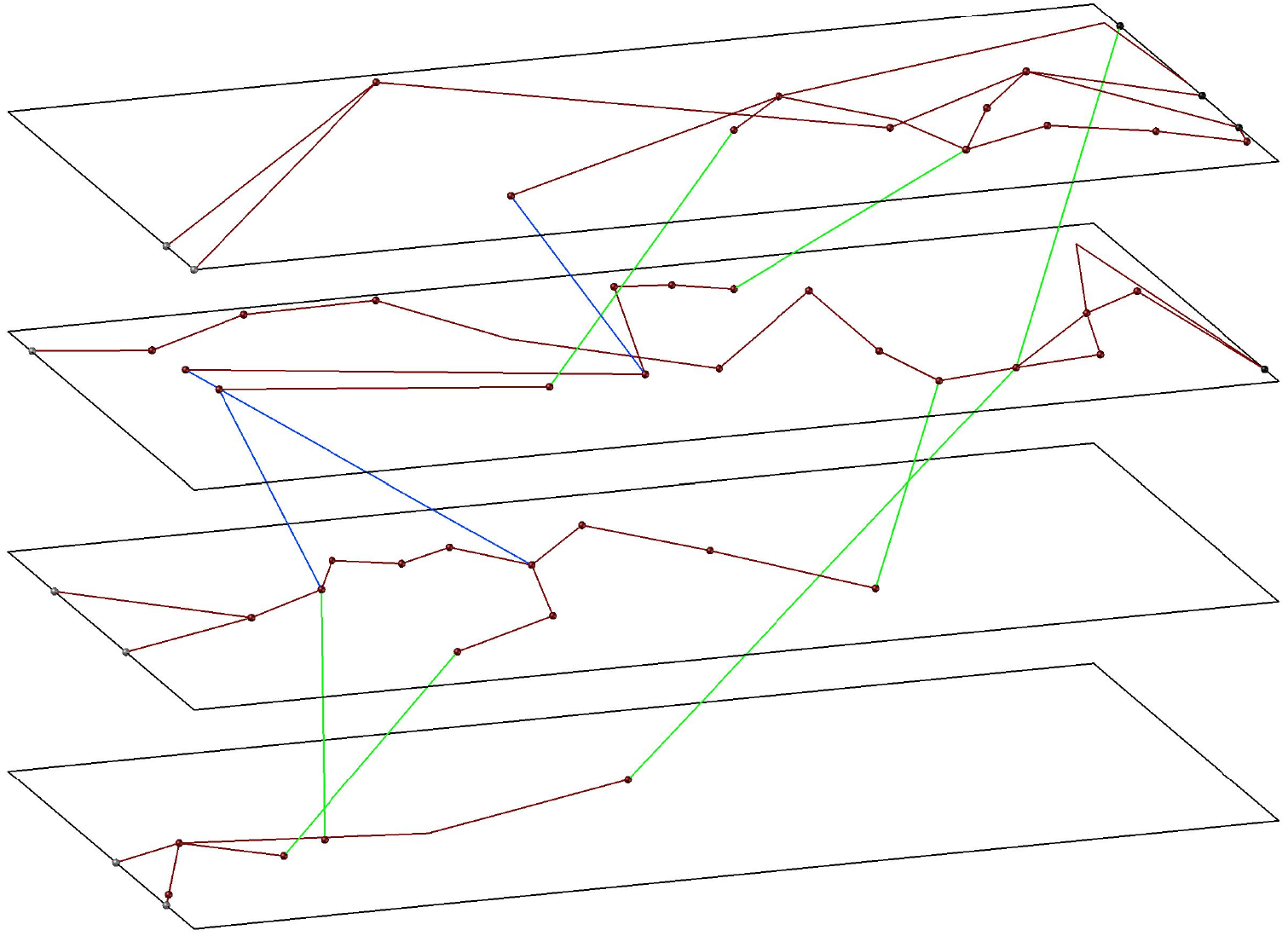} \label{fig:cants_move_11}} \\
    \end{tabular}
    \caption[]{{\bf Cant path selection and network construction (continued)}: (g) Several ants make their path from an input to an output. (h) The cants' nodes on each level are then condensed (clustered) based on their density using DBSCAN. \label{fig:cants_move}}
\end{figure}

\begin{figure}
    \ContinuedFloat 
    \centering 
    \begin{minipage}{0.4\textwidth}
    \subfloat[][]{\includegraphics[width=0.9\textwidth,height=0.2\textheight]{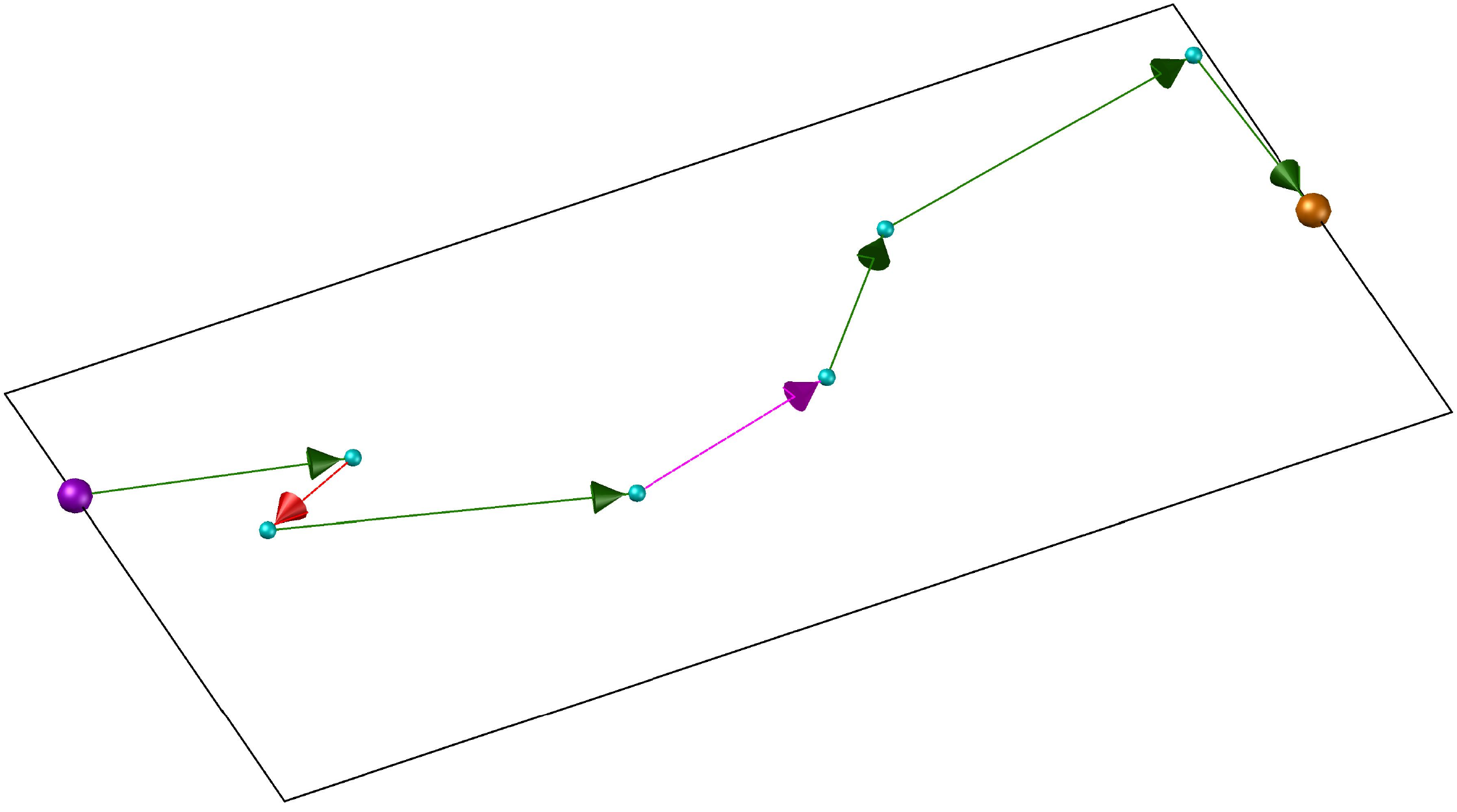} \label{fig:cants_move_08}} 
    \\
    \subfloat[][]{\includegraphics[width=0.9\textwidth,height=0.43\textheight]{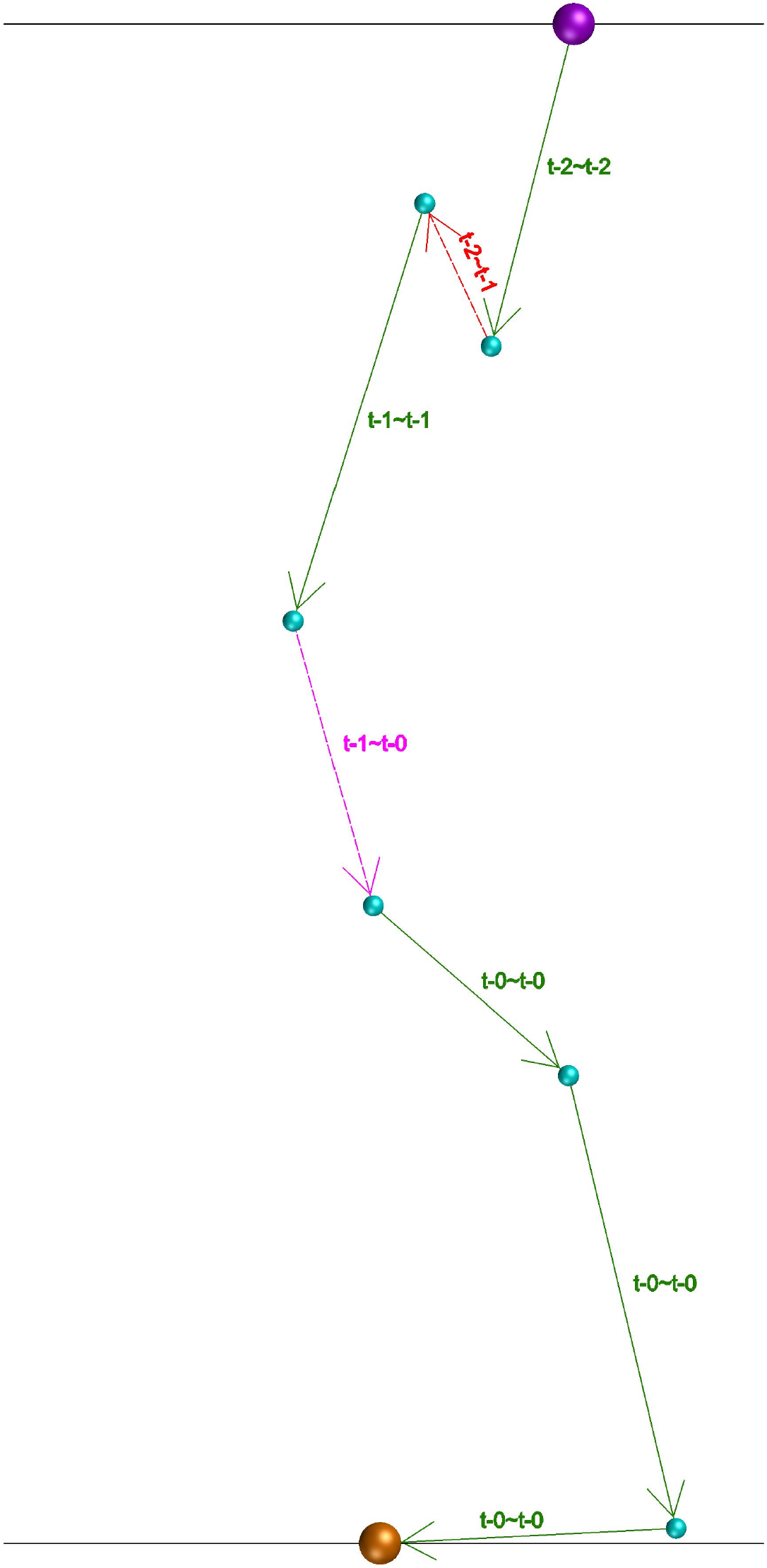} \label{fig:cants_move_09}}
    \end{minipage}%
    \begin{minipage}{0.4\textwidth}
            \subfloat[][]{\includegraphics[width=0.9\textwidth,height=0.67\textheight]{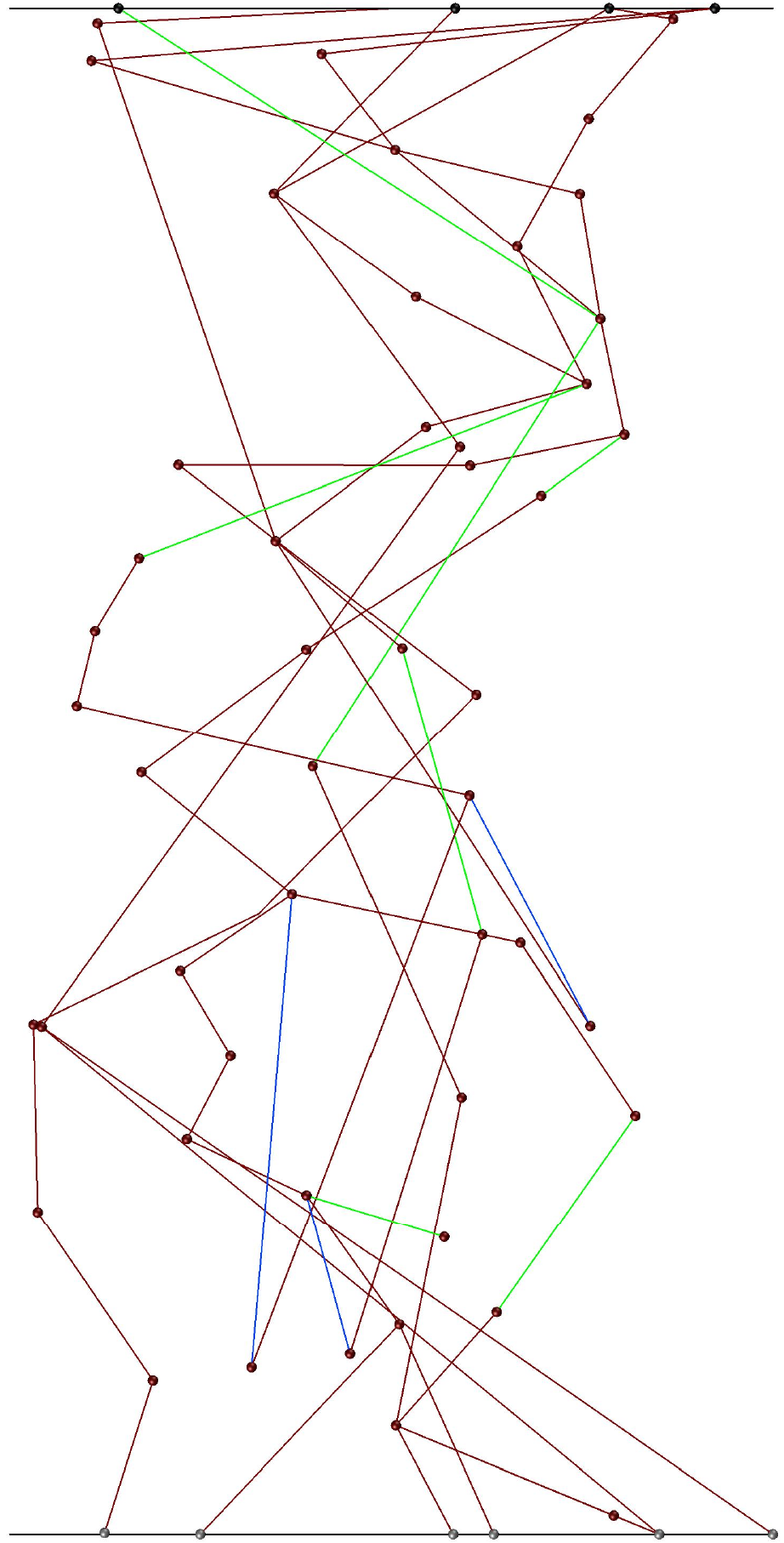} \label{fig:cants_move_12}}
    \end{minipage}
    \caption[]{{\bf Cant path selection and network construction (continued):} (i) An example cant's path is projected on one plane. (j) The cant picked its input point, starting at level $t_{-2}$, picked a node at $t_{-2}$ ({\color{ForestGreen} edge}), picked a node at $t_{-1}$ ({\color{red} backward recurrent edge}), picked a node at $t_{-1}$ ({\color{ForestGreen} edge}), picked a node at $t_{0}$ ({\color{magenta} forward recurrent edge}), picked a node at $t_0$ ({\color{ForestGreen} edge}), picked a node at $t_0$ ({\color{ForestGreen} edge}), and finally picked an output node at $t_{0}$ ({\color{ForestGreen} edge}). (k) The final network is the final result of clustering the nodes and defining the connections between nodes in the same layer as {\color{red}edges}, and the connection between nodes and between layers as {\color{green}forward recurrent edges} or {\color{blue}backward recurrent edges}. The flow moves from the {\color{gray} inputs} at the bottom to the {\color{black} outputs} at the top. \label{fig:cants_move2}}
\end{figure}

\paragraph{Cant Agent Input Node and Layer Selection:}
Each level in the search space has a level-selection pheromone value, $p_l$, where $l$ is the level. These are initialized to $p_l = 2 * l$ where the top level for the current time step is $l = 1$, the next level for the first time lag is $l = 2$ and so on. A cant selects its starting level according to the probability of starting at level $l$ as $P(l) = \frac{p_l}{\Sigma^L_{l=1} p_l}$, where $L$ is the total number of levels. This scheme encourages cants to start at lower levels of the stack at the beginning of the search. After selecting a level, the cant selects its input node in a similar fashion, based on the pheromones for each input node location on that level. When a candidate RNN is inserted into the population, the level pheromones for each level, utilized by that RNN, are incremented.

\paragraph{Cant Agent Movement:}
To balance exploration with exploitation, cants behave similarly to real-world ants by following communication clues to reach to targets. When a cant moves, it first decides if it will climb up to a higher (stack) level. This is done in the same manner as selecting its initial layer, except that it only selects between its current level and higher ones. After deciding if it will climb or not, the agent will then decide if it will explore or exploit. Cants randomly choose to exploit at a percentage equal to an exploitation parameter, $\epsilon$. 

When a cant decides to exploit and follow pheromone traces, \ie, clues, it will start sensing the pheromone points around it, given a sensing radius, $\rho$. If the cant is staying on the same level, it will only consider deposited pheromones that are in front of it (\ie, closer to the output nodes), otherwise, it will consider all the pheromones that are inside its sensing radius on the level it is moving to. The cant then calculates the center of mass of the pheromones in this region using the point in the space it will move to. This point is then saved by the candidate RNN (as a point to potentially increment pheromone values) if the RNN is later to be inserted into the RNN population. Since cants consider the center of mass of the pheromone values, the individual points of pheromone values are not the effective factor in the cant-to-cant communication. Rather, it is the concentration of the pheromone in a region of the space that more closely aligns with how real ants move in nature.

When a cant instead decides that it will explore, it instead selects a random point that lies within the range of their sensing radius to move to. Once a cant decides if it is climbing or staying in the same level, it will generate an angle bisector that is either a random number between $[0, 1]$ if the current and next point are on the same level or $[-1, 1]$ if the current and next points are on different levels. This angle bisector is used to calculate the angle of the next movement of the cant: $\theta=angle\_bisect * PI$. The movement angle is then subsequently used to calculate the next $x$ and $y$ coordinates of the next position of the cant: $x_{new} \gets x_{old} + \rho * cos(\theta)$, $y_{new} \gets y_{old} + \rho * sin(\theta)$. These points are also saved for potentially future pheromone modification.

\paragraph{Condensing Cant Paths to RNN Nodes:}
After cants choose the points in their paths from the inputs to the outputs, the points in the search space are clustered using the density-based spatial clustering of applications with noise~\cite{ester1996density} (DBSCAN) algorithm to condense those points to centroids. The points of the segments of the cants' paths are then shifted to the centroids that they belong to in the search space and those new points become the nodes of the generated RNN architecture (see Figures~\ref{fig:ants_multi_path} and~\ref{fig:ants_condensed_path}). The node types are picked by a pheromone-based discrete local search, as done in the discrete space ANTS. Each of these node types at the selected point will have their own pheromone values which drive probabilistic selection.

\paragraph{Communal Weight Sharing:}
\label{sec:cants_weight_inherite}
In order to avoid having to retrain every newly generated RNN from scratch, a communal weight sharing method has been implemented to allow generated RNNs to start with values similar to those of previously generated and trained RNNs. The centroid points (\ie, the RNN node points in the continuous space) in CANTS retain the weights of all the out-going edges from those nodes. Each newly-created centroid is assigned a weight value which is passed to the edges of the generated RNN. In case where a centroid did not have any previously created centroid in its cluster, randomly initialized weights are assigned to those outgoing edges either uniformly at random between $-0.5$ and $0.5$, or via the Kaiming~\cite{he2015delving}~or Xavier~\cite{glorot2010understanding}~strategies. If there were previously-created centroids in the clustering region, the weight values assigned to the generated RNN nodes are the average of the weights of those existing centroids. The weights of a centroid are updated after an RNN is trained by calculating the averages of the original centroid weight values and all the weights of the outgoing edges of the corresponding node (after training). The updated weights can then be used to initialize new centroid weights when they lie in their cluster when DBSCAN is applied in the following iteration. 

\paragraph{Pheromone Volatility:}
\label{sec:cants_pheromone_volatility}
Pheromone decay happens on a regular basis after each iteration of optimization regardless of the performance of the generated RNN(s). The pheromones decay by a constant value and after a specific minimum threshold the point is removed from the search space. By letting points vanish, the search space removes tiny residual pheromones which might provide distraction to cant-to-cant communication as well as slow down the overall algorithm.

\paragraph{Pheromone Incrementation:}
\label{sec:cants_pheromone_increment}
For each successful candidate RNN, \ie, each RNN that performs at least better than the worst in the population, the corresponding centroids for its RNN nodes in the search space are rewarded by increasing their pheromone values by a constant value. The values of the pheromones have a maximum limit to avoid becoming overly attractive points to the cants, which could result in premature convergence.

\section{Results}
\label{sec:results}

This work compares CANTS to the state-of-the-art ANTS and EXAMM algorithms on three real world datasets related to power systems. All three methods were used to perform time series data prediction for different parameters, which have been used as benchmarks in prior work. Main flame intensity was used as the prediction parameter from the coal plant's burner, net plant heat rate was used from the coal plant's boiler, and average power output was used from the wind turbines. Experiments were also performed to investigate the effect of CANTS hyper-parameters: the number of cants and cant sensing radii, $\epsilon$. 

\paragraph{\textbf{Computing Environment}}
\label{sec:computing}
The results for ANTS, CANTS and EXAMM were obtained by scheduling the experiment on {\bf redacted for double blind review}'s high performance computing cluster with $64$ Intel\textsuperscript{\textregistered} Xeon\textsuperscript{\textregistered} Gold $6150$ CPUs, each with 36 cores and $375$ GB RAM (total $2304$ cores and $24$ TB of RAM). Each ANTS experiment utilized $15$ nodes ($540$ cores), taking approximately $30$ days to complete all the experiments. CANTS experiments used $5$ nodes ($180$ cores), taking 7 days to finish all the experiments.  EXAMM experiments also used $5$ nodes ($180$ cores) and also took approximately 7 days to complete the experiments.

\paragraph{\textbf{Datasets}}
\label{sec:datasets} 
The datasets used, which are derived from coal-fired power plant and wind turbine data, have been previously made publicly available to encourage reproducibility\footnote{{\bf Redacted for double blind review.}}. The first dataset comes from measurements collected from $12$ burners of a coal-fired power plant as well as its boiler parameters and the second dataset comes from wind turbine engine data from the years $2013$ to $2020$, collected and made available by ENGIE's La Haute Borne open data windfarm\footnote{https://opendata-renewables.engie.com}. 

All of the datasets are multivariate and non-seasonal, with $12$ (burner), $48$ (boiler), and $78$ (wind turbine) input parameters (potentially dependent). These time series are very long, with the burner data separated into $7000$ time step chunks -- one for training and one for testing (per minute recordings). The boiler dataset is separated into a training set of $850$ steps and testing of $211$ steps (per hour recordings). The wind turbine dataset is separated into a training chunk of $190,974$ steps and  testing of $37,514$ steps (each step taken every $10$ minutes).

\begin{figure}
    \RawFloats
    \centering
    \begin{minipage}{0.49\textwidth}
        \centering
        \includegraphics[width=.9\textwidth]{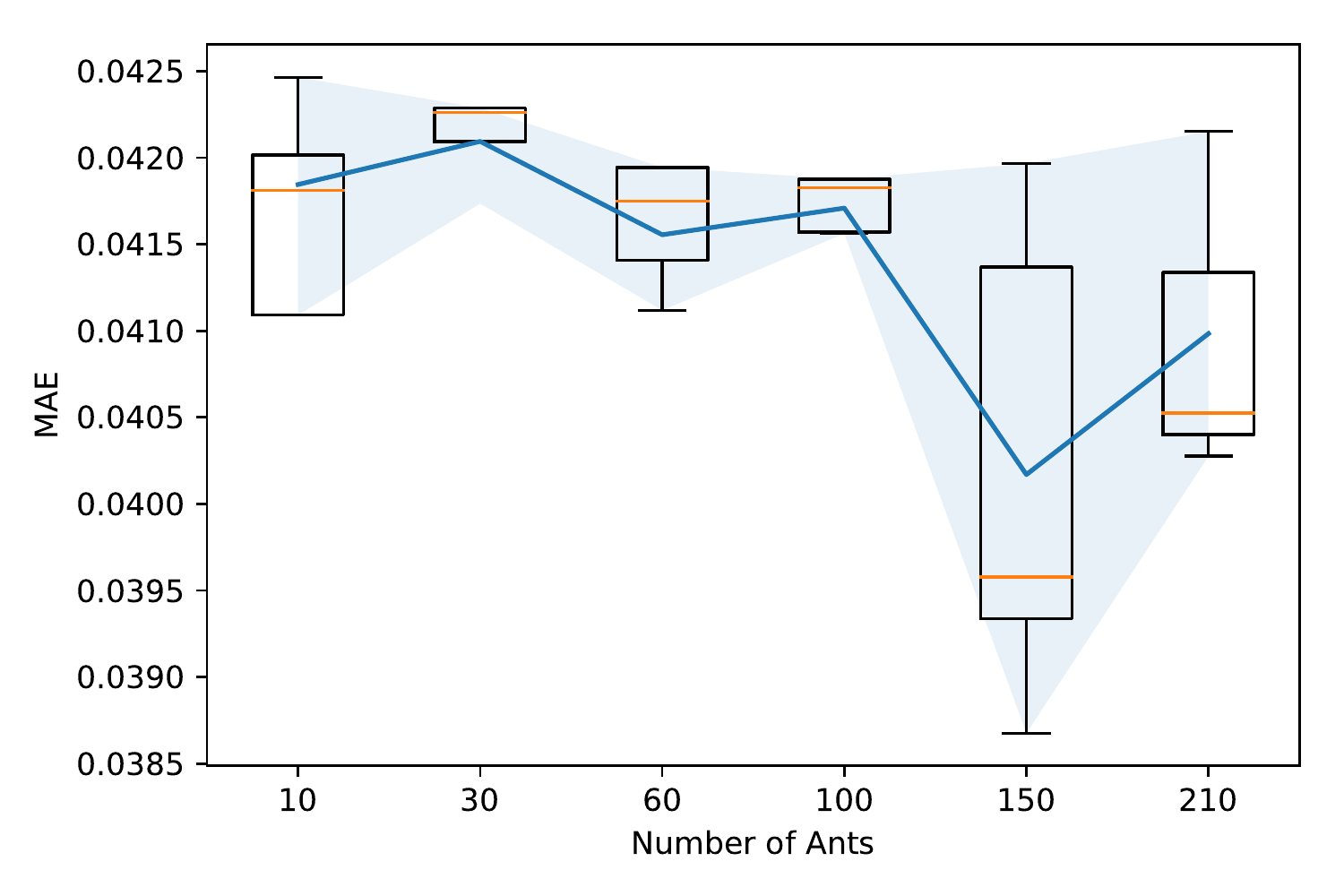}
        \caption{\label{fig:cants_num_ants} CANTS w/ varying \# of agents.}
    \end{minipage}
    \hfill
    \begin{minipage}{0.49\textwidth}
        \centering
        \includegraphics[width=.9\textwidth]{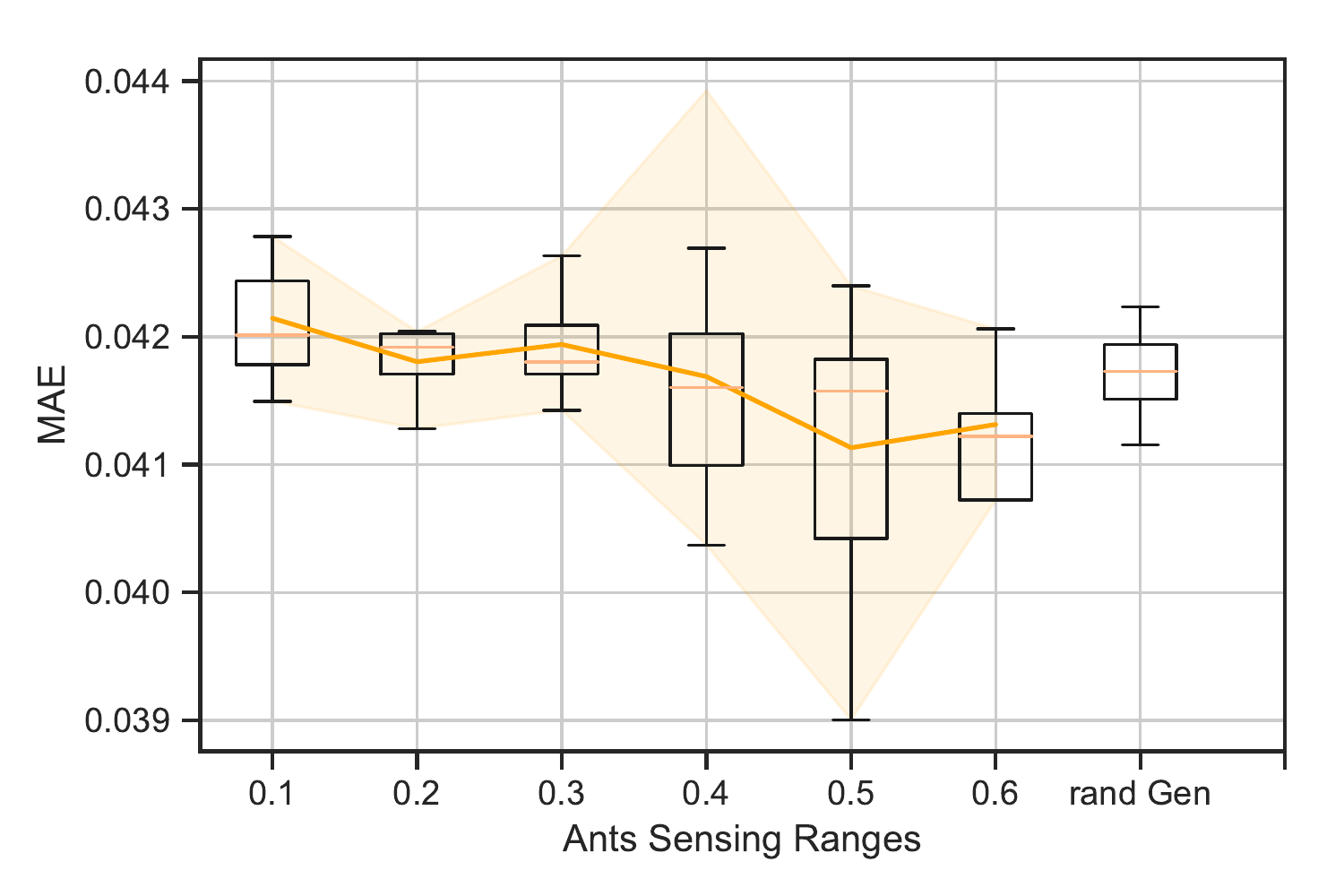}
        \caption{\label{fig:cants_radii} CANTS w/ different sensing radii.}
    \end{minipage}
\end{figure}

\subsection{Number of Cant Agents} 
An experiment was conducted to determine the effect that the number of cant agents has on the performance of CANTS. The experiment focusing on the net plant heat rate feature from the coal-fired power plant dataset. The number of ants evaluated were $10$, $30$, $60$, $100$, $150$, and $210$. The results, illustrated in Figure~\ref{fig:cants_num_ants}, show that, as the number of cants are increased, the performance increases until $150$ cants are used and then a decline is observed. This shows that the number of cant agents is an important parameter and requires tuning, potentially exhibiting ``sweet spots'' that, if uncovered, provide strong results. 

\subsection{Cant Agent Sensing Radius} 
We next investigated the effect that the sensing radii (range) of the cant agents had on algorithm performance.
Figure~\ref{fig:cants_radii} shows that a sensing radius of $0.5$ showed better performance compared to the $0.1$, $0.2$, $0.3$, $0.4$, and $0.6$ sensing radii values we tested. We also evaluated the effect that using a randomly generated sensing radius per cant agent would have.
For these, $\epsilon$ was randomly initialized (uniformly) via $\sim U(0.01, 0.98)$. Ultimately, we discovered that the sensing radius of $0.5$ still provided the best results.

\begin{figure}
    \RawFloats
    \centering
    \begin{minipage}[t]{0.49\textwidth}
        \centering
        \includegraphics[width=.95\textwidth,height=0.25\textheight]{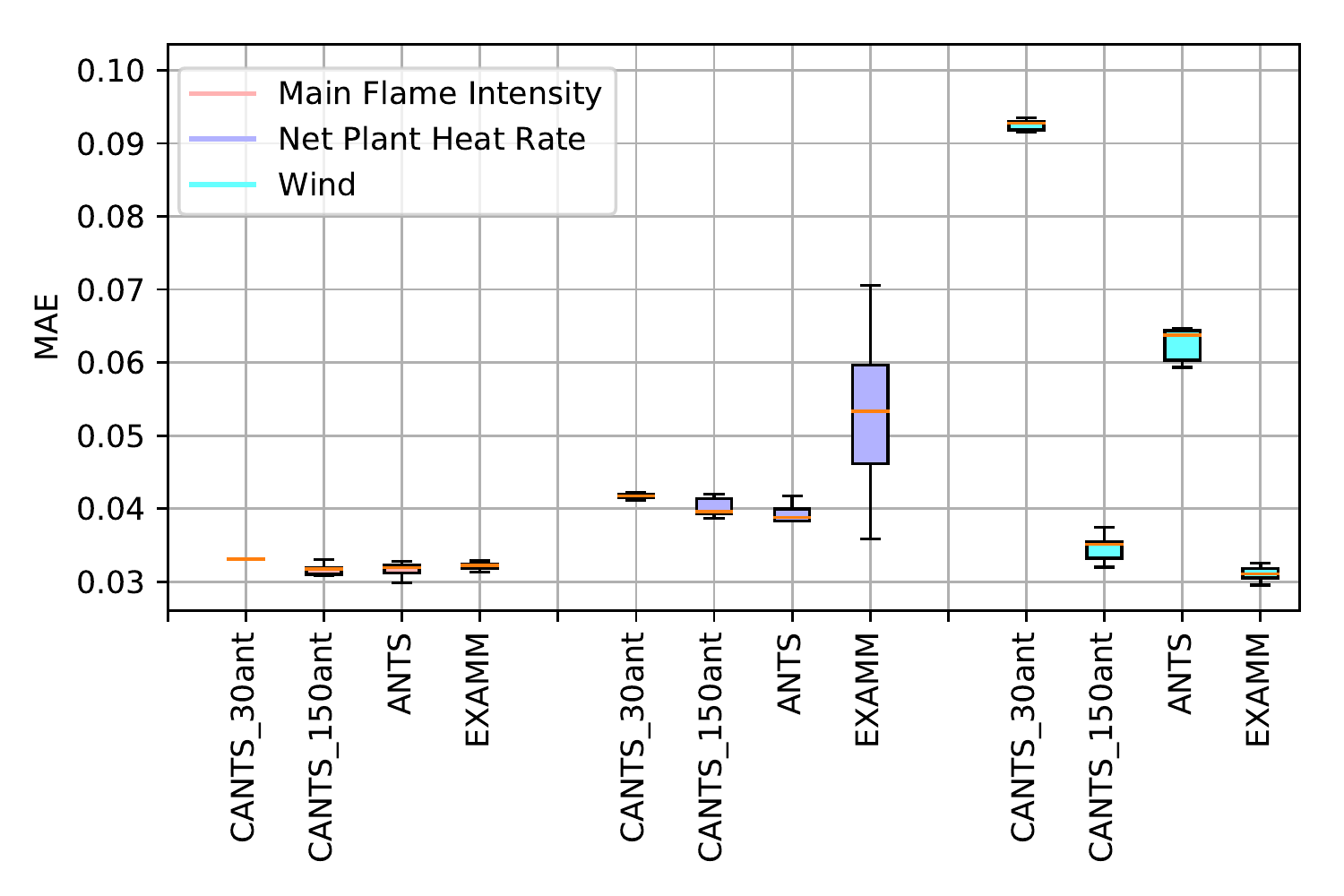}
        \caption{\label{fig_cants_maes} Mean Average Error (MAE) ranges of best-found RNNs from each method.}
    \end{minipage}
    \hfill
    \begin{minipage}[t]{0.49\textwidth}
    \centering
    \includegraphics[width=.95\textwidth,height=0.25\textheight]{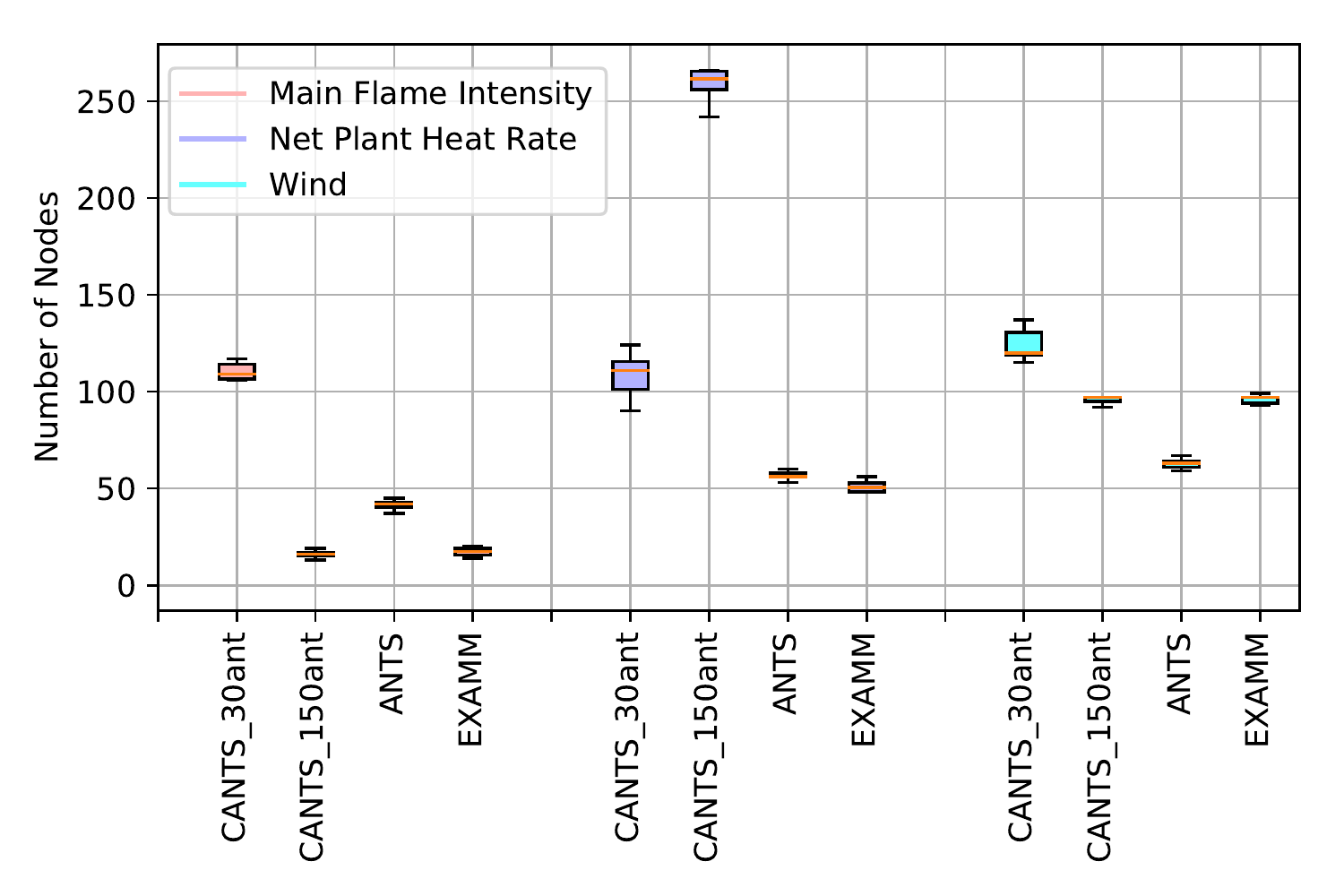}
    \caption{\label{fig_cants_nodes} Number of nodes in the best found RNNs from each method.}
    \end{minipage}
\end{figure}

\subsection{Algorithm Benchmark Comparisons}
To compare the three different NAS strategies, each experiment was repeated $10$ times (trials) for statistical comparison and all algorithms were set to generate $2000$ RNNs per trial. For CANTS, the sensing radii of the cant agents and exploration instinct values were generated uniformly via $\sim U(0.01, 0.98)$ when the cants were created, initial pheromone values were $1$ and the maximum was kept at $10$ with a pheromone decay rate set to $0.05$. For the DBSCAN module, clustering distance was $0.05$ with a minimum point value of $2$ -- runs with these settings were done using $30$ and $150$ ants. CANTS and ANTS used a population of size $20$ while EXAMM used $4$ islands, each with a population of $10$. ANTS, CANTS, and EXAMM all had a maximum recurrent depth of $5$ and the predictions were made over a forecasting horizon of $1$. The generated RNNs were each allowed $40$ epochs of back-propagation for local fine-tuning (since all algorithms are mmetic). ANTS and EXAMM utilized the hyper-parameters previously reported to yield best results~\cite{ororbia2019examm,elsaid2020ant}.

\begin{figure}
    \RawFloats
    \begin{minipage}[t]{0.49\textwidth}
    \centering
    \includegraphics[width=.9\textwidth,height=0.25\textheight]{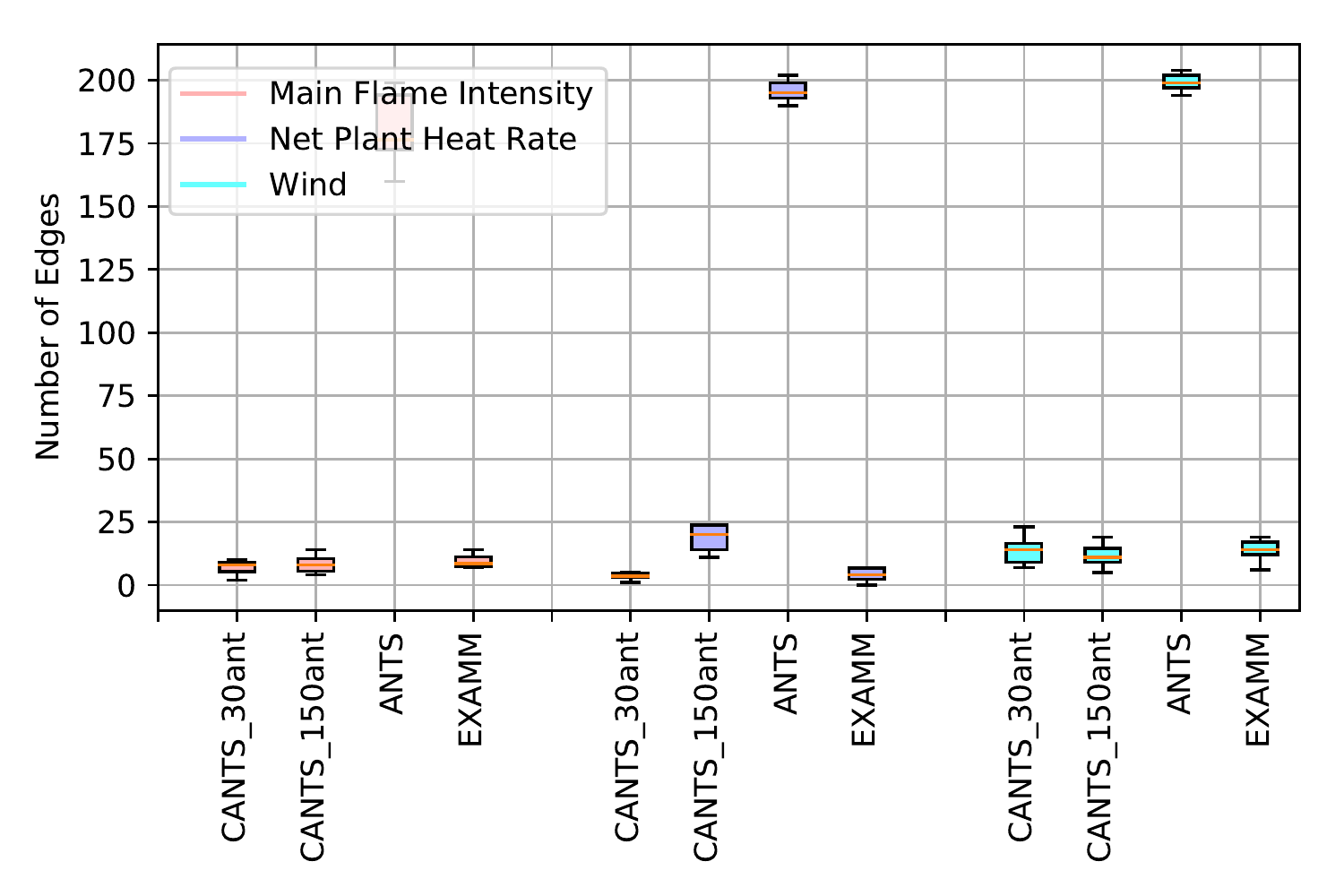}
    \caption{\label{fig_cants_edges} Number of edges in the best-found RNNs from each algorithm.}
    \end{minipage}
    \hfill
    \begin{minipage}[t]{0.49\textwidth}
    \centering
    \includegraphics[width=.95\textwidth,height=0.25\textheight]{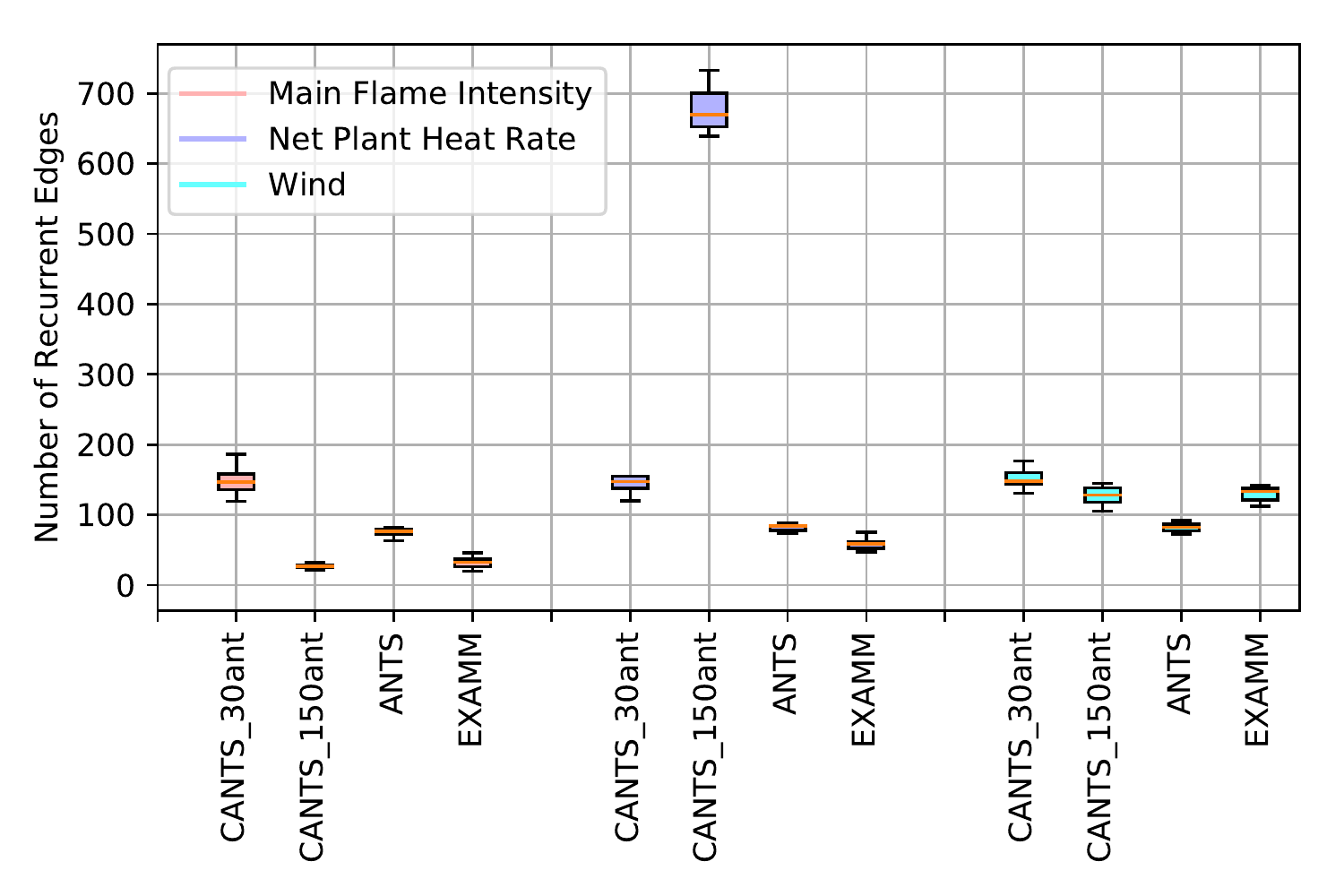}
    \caption{\label{fig_cants_recedges} Number of recurrent edges in the best-found RNNs each algorithm. }
    \end{minipage}
\end{figure}

The results shown in Figure~\ref{fig_cants_maes}, which compare CANTS, ANTS and EXAMM in the three experiments described above over the three datasets, report the range of mean average error (MAE) of each algorithm's best-found RNNs. While EXAMM outperformed CANTS with $30$ ants, CANTS with $150$ ants had a better performance than EXAMM and ANTS. CANTS was competitive with ANTS on the net plant heat rate predictions and outperformed EXAMM on this dataset. CANTS also outperformed ANTS on the wind energy dataset yet could not beat EXAMM. Potential reasons for this could be that the complexity/size of this dataset is greater and that the task is simply more difficult which results in a potentially larger search space. As CANTS allows for potentially unbounded network sizes, its search space is significantly larger than either that of ANTS or EXANM. Though ANTS outperformed CANTS on the wind dataset, CANTS is still a good competitor, especially since it has less hyper-parameters ($8$) to tune compared to both ANTS and EXAMM (both require at least $16$).  While all these reasons may be valid, the size of the search space is likely the biggest challenge. Further evidence of this is provided in Figures~\ref{fig_cants_nodes}, \ref{fig_cants_edges}, \ref{fig_cants_recedges}, present the number of structural elements (nodes, edges, and recurrent edges, respectively) of the best-found RNN architectures using the different algorithms. The CANTS runs with $150$ ants resulted in significantly more complex architectures for many of the problems, which may be an indication that CANTS can evolve better performing structure if provided more optimization iterations.

\section{Discussion and Future Work}
\label{sec:conclusion}

This work introduces continuous ant-based neural topology search (CANTS), a novel nature-inspired optimization algorithm that utilizes a continuous search space to conduct unbounded neural architecture search (NAS). This approach provides a unique strategy to overcome key limitations of constructive neuro-evolutionary strategies (which often prematurely get stuck at finding smaller, less performant architectures) as well as other neural architecture search strategies that require users to carefully specify the bounds limiting the neural architecture sizes. CANTS was experimentally evaluated for the automated design of recurrent neural networks (RNNs) to make time series predictions across three challenging real-world data sets in the power systems domain. We compared it to two state-of-the-art algorithms, ANTS (a discrete space ant colony NAS algorithm) and EXAMM (a constructive neuro-evolution algorithm). CANTS is shown to improve on or be competitive with these strategies, while also being simpler to use and tune, only requiring $8$ hyper-parameters as opposed to the $16$ hyper-parameters of the other two strategies.

This study presents some initial work generalizing ant colony algorithms to complex, continuous search spaces, specifically for unbounded graph optimization problems (with NAS as a target application), opening up a number of promising avenues for future work. In particular, while the search space is continuous in each two-dimensional slice (or time step) of our temporal stack, there is still the number of discrete levels that a user must specify. Therefore, a promising extension of the algorithm would be to make the search space continuous across all three dimensions, removing this parameter, and allowing pheromone placements to guide the depth of recurrent connections. This could have implications for discrete-event, continuous-time RNN models \cite{mozer2017discrete}, which attempt to tackle a broader, more interesting set of sequence modeling problems.
Finally, and potentially the most interesting, is the fact that the exploitation parameter, $\epsilon$, and the sensing radius, $\rho$, for each synthetic ant agent in our algorithm was held fixed (or in some cases randomly initialized) for the duration of each CANTS search. However, the ants could instead be treated as complex agents that evolve with time, learning the best exploitation and sensing parameters for the task search spaces they are applied to. This could provide far greater flexibility to the CANTS framework. Expanding this algorithm to other domains, such as the automated design of convolutional neural networks (for computer vision) or to other types of RNNs, such as those used for natural language processing, could further demonstrate the potentially broad applicability of this nature-inspired approach.

\section*{Acknowledgements}
This material is in part supported by the U.S. Department of Energy, Office of Science, Office of Advanced Combustion Systems under Award Number \#FE0031547. We also thank Microbeam Technologies, Inc. for their help in collecting and preparing the coal-fired power plant dataset.
Most of the computation of this research was done on the high performance computing clusters of Research Computing at Rochester Institute of Technology. We would like to thank the Research Computing team for their assistance and the support they generously offered to ensure that the heavy computation this study required was available.

\bibliographystyle{unsrt} 
\bibliography{bibliography_merged.bib}

\begin{thebibliography}{10}

\bibitem{zoph2016neural}
Barret Zoph and Quoc~V Le.
\newblock Neural architecture search with reinforcement learning.
\newblock {\em arXiv preprint arXiv:1611.01578}, 2016.

\bibitem{erkaymaz2014impact}
Okan Erkaymaz, Mahmut {\"O}zer, and Nejat Yumu{\c{s}}ak.
\newblock Impact of small-world topology on the performance of a feed-forward
  artificial neural network based on 2 different real-life problems.
\newblock {\em Turkish Journal of Electrical Engineering \& Computer Sciences},
  22(3):708--718, 2014.

\bibitem{Barna1990}
Gy{\"o}rgy Barna and Kimmo Kaski.
\newblock {\em Choosing optimal network structure}, pages 890--893.
\newblock Springer Netherlands, Dordrecht, 1990.

\bibitem{elsken2018neural}
Thomas Elsken, Jan~Hendrik Metzen, and Frank Hutter.
\newblock Neural architecture search: A survey.
\newblock {\em arXiv preprint arXiv:1808.05377}, 2018.

\bibitem{liu2018darts}
Hanxiao Liu, Karen Simonyan, and Yiming Yang.
\newblock Darts: Differentiable architecture search.
\newblock {\em arXiv preprint arXiv:1806.09055}, 2018.

\bibitem{pham2018efficient}
Hieu Pham, Melody~Y Guan, Barret Zoph, Quoc~V Le, and Jeff Dean.
\newblock Efficient neural architecture search via parameter sharing.
\newblock {\em arXiv preprint arXiv:1802.03268}, 2018.

\bibitem{xie2018snas}
Sirui Xie, Hehui Zheng, Chunxiao Liu, and Liang Lin.
\newblock Snas: stochastic neural architecture search.
\newblock {\em arXiv preprint arXiv:1812.09926}, 2018.

\bibitem{luo2018neural}
Renqian Luo, Fei Tian, Tao Qin, Enhong Chen, and Tie-Yan Liu.
\newblock Neural architecture optimization.
\newblock In {\em Advances in neural information processing systems}, pages
  7816--7827, 2018.

\bibitem{stanley2019designing}
Kenneth~O Stanley, Jeff Clune, Joel Lehman, and Risto Miikkulainen.
\newblock Designing neural networks through neuroevolution.
\newblock {\em Nature Machine Intelligence}, 1(1):24--35, 2019.

\bibitem{darwish2020survey}
Ashraf Darwish, Aboul~Ella Hassanien, and Swagatam Das.
\newblock A survey of swarm and evolutionary computing approaches for deep
  learning.
\newblock {\em Artificial Intelligence Review}, 53(3):1767--1812, 2020.

\bibitem{horng2017fine}
Ming-Huwi Horng.
\newblock Fine-tuning parameters of deep belief networks using artificial bee
  colony algorithm.
\newblock {\em DEStech Transactions on Computer Science and Engineering}, 2017.

\bibitem{yang2010new}
Xin-She Yang.
\newblock A new metaheuristic bat-inspired algorithm.
\newblock In {\em Nature inspired cooperative strategies for optimization
  (NICSO 2010)}, pages 65--74. Springer, 2010.

\bibitem{yang2010nature}
Xin-She Yang.
\newblock {\em Nature-inspired metaheuristic algorithms}.
\newblock Luniver press, 2010.

\bibitem{leke2017deep}
Collins Leke, Alain~Richard Ndjiongue, Bhekisipho Twala, and Tshilidzi Marwala.
\newblock A deep learning-cuckoo search method for missing data estimation in
  high-dimensional datasets.
\newblock In {\em International Conference on Swarm Intelligence}, pages
  561--572. Springer, 2017.

\bibitem{dorigo1996ant}
Marco Dorigo, Vittorio Maniezzo, and Alberto Colorni.
\newblock Ant system: optimization by a colony of cooperating agents.
\newblock {\em IEEE Transactions on Systems, Man, and Cybernetics, Part B
  (Cybernetics)}, 26(1):29--41, 1996.

\bibitem{desell2015evolving}
Travis Desell, Sophine Clachar, James Higgins, and Brandon Wild.
\newblock Evolving deep recurrent neural networks using ant colony
  optimization.
\newblock In Gabriela Ochoa and Francisco Chicano, editors, {\em Evolutionary
  Computation in Combinatorial Optimization}, pages 86--98, Cham, 2015.
  Springer International Publishing.

\bibitem{mavrovouniotis2013evolving}
Michalis Mavrovouniotis and Shengxiang Yang.
\newblock Evolving neural networks using ant colony optimization with pheromone
  trail limits.
\newblock In {\em Computational Intelligence (UKCI), 2013 13th UK Workshop on},
  pages 16--23. IEEE, 2013.

\bibitem{elsaid2018optimizing}
AbdElRahman ElSaid, Fatima El~Jamiy, James Higgins, Brandon Wild, and Travis
  Desell.
\newblock Optimizing long short-term memory recurrent neural networks using ant
  colony optimization to predict turbine engine vibration.
\newblock {\em Applied Soft Computing}, 73:969--991, 2018.

\bibitem{elsaid2020ant}
AbdElRahman ElSaid, Alexander~G Ororbia, and Travis~J Desell.
\newblock Ant-based neural topology search (ants) for optimizing recurrent
  networks.
\newblock In {\em International Conference on the Applications of Evolutionary
  Computation (Part of EvoStar)}, pages 626--641. Springer, 2020.

\bibitem{cai2018proxylessnas}
Han Cai, Ligeng Zhu, and Song Han.
\newblock Proxylessnas: Direct neural architecture search on target task and
  hardware.
\newblock {\em arXiv preprint arXiv:1812.00332}, 2018.

\bibitem{guo2020single}
Zichao Guo, Xiangyu Zhang, Haoyuan Mu, Wen Heng, Zechun Liu, Yichen Wei, and
  Jian Sun.
\newblock Single path one-shot neural architecture search with uniform
  sampling.
\newblock In {\em European Conference on Computer Vision}, pages 544--560.
  Springer, 2020.

\bibitem{bender2018understanding}
Gabriel Bender, Pieter-Jan Kindermans, Barret Zoph, Vijay Vasudevan, and Quoc
  Le.
\newblock Understanding and simplifying one-shot architecture search.
\newblock In {\em International Conference on Machine Learning}, pages
  550--559, 2018.

\bibitem{dong2019one}
Xuanyi Dong and Yi~Yang.
\newblock One-shot neural architecture search via self-evaluated template
  network.
\newblock In {\em Proceedings of the IEEE International Conference on Computer
  Vision}, pages 3681--3690, 2019.

\bibitem{zhao2020few}
Yiyang Zhao, Linnan Wang, Yuandong Tian, Rodrigo Fonseca, and Tian Guo.
\newblock Few-shot neural architecture search.
\newblock {\em arXiv preprint arXiv:2006.06863}, 2020.

\bibitem{stanley2002evolving}
Kenneth~O Stanley and Risto Miikkulainen.
\newblock Evolving neural networks through augmenting topologies.
\newblock {\em Evolutionary computation}, 10(2):99--127, 2002.

\bibitem{ororbia2019examm}
Alexander Ororbia, AbdElRahman ElSaid, and Travis Desell.
\newblock Investigating recurrent neural network memory structures using
  neuro-evolution.
\newblock In {\em Proceedings of the Genetic and Evolutionary Computation
  Conference}, GECCO '19, pages 446--455, New York, NY, USA, 2019. ACM.

\bibitem{stanley2009hypercube}
Kenneth~O Stanley, David~B D'Ambrosio, and Jason Gauci.
\newblock A hypercube-based encoding for evolving large-scale neural networks.
\newblock {\em Artificial life}, 15(2):185--212, 2009.

\bibitem{ororbia2017diff}
Alexander~G. Ororbia~II, Tomas Mikolov, and David Reitter.
\newblock Learning simpler language models with the differential state
  framework.
\newblock {\em Neural Computation}, 0(0):1--26, 2017.
\newblock PMID: 28957029.

\bibitem{chung2014empirical}
Junyoung Chung, Caglar Gulcehre, KyungHyun Cho, and Yoshua Bengio.
\newblock Empirical evaluation of gated recurrent neural networks on sequence
  modeling.
\newblock {\em arXiv preprint arXiv:1412.3555}, 2014.

\bibitem{hochreiter1997long}
Sepp Hochreiter and J{\"u}rgen Schmidhuber.
\newblock Long short-term memory.
\newblock {\em Neural computation}, 9(8):1735--1780, 1997.

\bibitem{zhou2016minimal}
Guo-Bing Zhou, Jianxin Wu, Chen-Lin Zhang, and Zhi-Hua Zhou.
\newblock Minimal gated unit for recurrent neural networks.
\newblock {\em International Journal of Automation and Computing},
  13(3):226--234, 2016.

\bibitem{collins2016capacity}
Jasmine Collins, Jascha Sohl-Dickstein, and David Sussillo.
\newblock Capacity and trainability in recurrent neural networks.
\newblock {\em arXiv preprint arXiv:1611.09913}, 2016.

\bibitem{socha2008ant}
Krzysztof Socha and Marco Dorigo.
\newblock Ant colony optimization for continuous domains.
\newblock {\em European journal of operational research}, 185(3):1155--1173,
  2008.

\bibitem{kuhn2002ant}
Lachlan~D Kuhn.
\newblock Ant colony optimization for continuous spaces.
\newblock {\em Computer Science and Computer Engineering Undergraduate Honors
  Theses (35)}, 2002.

\bibitem{xiao2011hybrid}
Jing Xiao and LiangPing Li.
\newblock A hybrid ant colony optimization for continuous domains.
\newblock {\em Expert Systems with Applications}, 38(9):11072--11077, 2011.

\bibitem{gupta2014transistor}
Himanshu Gupta and Bahniman Ghosh.
\newblock Transistor size optimization in digital circuits using ant colony
  optimization for continuous domain.
\newblock {\em International Journal of Circuit Theory and Applications},
  42(6):642--658, 2014.

\bibitem{bilchev1995ant}
George Bilchev and Ian~C Parmee.
\newblock The ant colony metaphor for searching continuous design spaces.
\newblock In {\em AISB workshop on evolutionary computing}, pages 25--39.
  Springer, 1995.

\bibitem{ester1996density}
Martin Ester, Hans-Peter Kriegel, J{\"o}rg Sander, Xiaowei Xu, et~al.
\newblock A density-based algorithm for discovering clusters in large spatial
  databases with noise.
\newblock In {\em Kdd}, volume~96, pages 226--231, 1996.

\bibitem{he2015delving}
Kaiming He, Xiangyu Zhang, Shaoqing Ren, and Jian Sun.
\newblock Delving deep into rectifiers: Surpassing human-level performance on
  imagenet classification.
\newblock In {\em Proceedings of the IEEE international conference on computer
  vision}, pages 1026--1034, 2015.

\bibitem{glorot2010understanding}
Xavier Glorot and Yoshua Bengio.
\newblock Understanding the difficulty of training deep feedforward neural
  networks.
\newblock In {\em Proceedings of the thirteenth international conference on
  artificial intelligence and statistics}, pages 249--256, 2010.

\bibitem{mozer2017discrete}
Michael~C Mozer, Denis Kazakov, and Robert~V Lindsey.
\newblock Discrete event, continuous time rnns.
\newblock {\em arXiv preprint arXiv:1710.04110}, 2017.

\end{thebibliography}

\end{document}